\documentclass{article}

\PassOptionsToPackage{numbers, compress}{natbib}

 \usepackage[preprint]{neurips_2026}
\usepackage{wrapfig2}
\usepackage[utf8]{inputenc} 
\usepackage[T1]{fontenc}    
\usepackage{hyperref}       
\usepackage{url}            
\usepackage{graphicx}       
\usepackage{booktabs}       
\usepackage{amsmath}        
\usepackage{amsfonts}       
\usepackage{nicefrac}       
\usepackage{microtype}      
\usepackage{xcolor}         
\usepackage{natbib}
\usepackage{enumitem}
\usepackage{amsmath, amssymb, amsthm}
\usepackage{framed} 
\newtheorem{definition}{Definition}[section]  

\usepackage{xcolor}
\usepackage{amssymb}

\usepackage[most]{tcolorbox} 
\newtcolorbox{promptbox}[1][]{
  colback=gray!5,
  colframe=gray!50,
  fonttitle=\bfseries,
  breakable,
  enhanced,
  left=4pt, right=4pt, top=4pt, bottom=4pt,
  segmentation style={solid, gray!50},
  #1
}

\usepackage{multirow}
\usepackage{xcolor}
\usepackage{array}
\newcolumntype{x}[1]{>{\centering\arraybackslash}p{#1pt}}

\newcommand{\pos}[1]{\textcolor{green!50!black}{\scriptsize{(+#1)}}}
\newcommand{\nega}[1]{\textcolor{red!70!black}{\scriptsize{(-#1)}}}
\usepackage{makecell}

\title{The World Won't Stay Still: \\
  Programmable Evolution for Agent Benchmarks}

\author{
  \textbf{Guangrui Li\textsuperscript{1}} \quad
  \textbf{Yaochen Xie\textsuperscript{1}\footnotemark[1]\thanks{Equal contribution.}} \quad
  \textbf{Yi Liu\textsuperscript{1}\footnotemark[1]} \quad
  \textbf{Ziwei Dong\textsuperscript{1}\footnotemark[1]} \quad
  \textbf{Xingyuan Pan\textsuperscript{1}\footnotemark[1]} \quad \\ [0.3em] 
  \textbf{Tianqi Zheng\textsuperscript{1}} \quad
  \textbf{Jason Choi\textsuperscript{1}} 
  \textbf{Michael Morais\textsuperscript{1}} \quad
  \textbf{Binit Jha\textsuperscript{1}} \quad
  \textbf{Shaunak Mishra\textsuperscript{1}} \quad \\ [0.3em] 
  \textbf{Bingrou Zhou\textsuperscript{1}} \quad
  \textbf{Chen Luo\textsuperscript{1}} \quad
  \textbf{Monica Cheng\textsuperscript{1}} \quad
  \textbf{Dawn Song\textsuperscript{2}} \\ [0.5em]
  \textsuperscript{1}Amazon \quad
  \textsuperscript{2}UC Berkeley
}

\begin{document}

\maketitle

\begin{abstract}

LLM-powered tool-calling agents fulfill user requests by interacting with environments, querying data, and invoking tools in a multi-turn process. 
Yet most existing benchmarks evaluate these systems under static environment interfaces with fixed schemas and toolsets, making it difficult to assess how agents behave as environments evolve --- when capabilities are added, reorganized, or deprecated across successive environment versions.
In this paper, we study \emph{structured environment evolution} as a benchmark-construction problem for tool-calling agents.
We propose \textsc{ProEvolve}, a graph-based framework that makes environment evolution \emph{programmable}. At its core, a typed relational graph provides a unified, explicit representation of the environment --- data, tools, and schemas --- and their dependencies. Under this formalism, adding, removing, or modifying capabilities is expressed as graph transformations that coherently propagate updates across tools, schemas, and data access. Building on this, \textsc{ProEvolve} supports
(1) automatic generation of evolved executable environments through explicit graph transformations, and (2) graph-grounded construction of task sandboxes via subgraph sampling and instantiation.
We validate \textsc{ProEvolve} in two tool-calling domains, e-commerce and airline booking, in terms of quality, implementation validity, and failure modes.
Finally, we use the generated benchmark as a downstream diagnostic to study how representative agents behave under structured environment evolution.  
\end{abstract}

\section{Introduction}
Tool-calling agents operate over structured environments in which schemas, data entities, and tool interfaces jointly determine what information can be accessed and what actions can be executed~\citep{yao2023react,barres2025tau2,ALFRED20,yang2018hotpotqa}. Despite recent progress, existing approaches typically evaluate agents under \textit{static} environments, characterized by fixed toolsets and fixed  data schemas that specify accessible fields and entities in a deterministic fashion \citep{liu2023agentbench,qin2023toolllm,mialon2023gaia, zhou2023webarena,jimenez2023swe}. This assumption simplifies benchmark design, but it limits our ability to assess how agents behave when capabilities are added, reorganized, or deprecated across environment versions.

Recent efforts have expanded the scale and diversity of agent benchmarks by increasing the number of tools or tasks~\citep{li2024autobencher, shi2025taskcraft, yao2024tau, chen2024learning}, or by introducing curated sets of distinct environments~\citep{zhang2025autoenv}. These approaches improve coverage, but they typically treat environments as fixed snapshots or independently generated variants, rather than as successive versions linked by structured transitions. As a result, they capture environment variation, but not structured environment evolution.
In tool-calling environments, however, tools, data entities, and schemas are tightly coupled: adding, reorganizing, or deprecating a capability often requires coordinated updates across multiple components. For example, adding an order-cancellation capability to an order-placement system may reuse and extend existing order schemas, data entities, and tool interfaces, rather than introducing an isolated subsystem. Thus, what distinguishes environment \emph{evolution} from environment \emph{variation} is version linkage: each version is derived from its predecessor, unmodified structure is preserved, and intended changes are propagated coherently across coupled components.

In this paper, we study structured environment evolution as a benchmark-construction problem for tool-calling agents. We focus on environments where schemas, data entities, and tools are tightly coupled, such that modifying one component often requires coordinated updates to others. We argue that environment evolution is distinct from task scaling, independent environment variation, and surface-level tool perturbation: an evolved environment should be derived from a previous version, preserve unmodified structure, and propagate intended changes coherently across schemas, tools, data access, executable behavior, and generated tasks. This leads to two desiderata: (1) scalable coherence --- generating diverse environment versions without breaking dependencies, and (2) controlled versioned dynamics --- making each transition explicit, reproducible, and evaluable.



To address this problem, we propose \textsc{ProEvolve}, a graph-based framework for programming structured evolution in tool-calling environments. \textsc{ProEvolve} represents each environment as a typed relational graph over schemas, data entities, and tool-enabled transitions. Under this formalism, semantic evolution intents such as capability addition, access-path reorganization, and deprecation are implemented as explicit graph transformations. The transformed graph is then instantiated into executable environments and graph-grounded task sandboxes, with validation of implementation correctness, semantic intent alignment, diversity, and generation failure modes. We evaluate \textsc{ProEvolve} in two structured tool-calling domains, e-commerce and airline booking, and use downstream agent results as diagnostic evidence that the generated environment changes are meaningful.

Concretely, \textsc{ProEvolve} produces 200 environments versions and 3{,}000 task sandboxes in e-commerce and 20 versions  with 300 sandboxes in airline booking. Beyond validating implementation correctness, semantic fidelity, and evolution diversity, we demonstrate two novel uses enabled by version-linked benchmarks. First, \emph{counterfactual capability
measurement}, where we isolate agent scalability, efficiency, and robustness as separate, quantifiable dimensions, which is impossible with existing independently sampled environments. Second, \emph{memory-based
adaptation}, where we show that memory strategies developed for
static settings often reduce tool usage but degrade task completion
under structured change, revealing that prior-version experience can
become stale or misleading when tools and schemas evolve.

\section{Related Work}

Existing agent evaluation paradigms predominantly assess agents in \textit{static} environments with fixed schemas and toolsets. These benchmarks are based on environments that are snapshots of real-world applications or constructed via substantial human effort. SWE-bench \citep{jimenez2023swe, yang2024swe} and LiveCodeBench \citep{jain2024livecodebench} are widely adopted to evaluate software engineering agents, while OSWorld \citep{xie2024osworld} and WebArena \cite{zhou2023webarena} evaluate multimodal control in fixed operating systems and web environments. Generalist benchmarks such as AgentBench \cite{liu2023agentbench}, ToolBench \cite{qin2023toolllm}, and GAIA \cite{mialon2023gaia} assess capabilities in diverse, heterogeneous domains—ranging from database management to household tasks.

To address scalability and diversity, recent work has shifted toward synthetically generating tasks~\cite{xie2025agentsynth} and environments. Systems like AutoBencher \cite{li2024autobencher} and TaskCraft \cite{shi2025taskcraft} keep the underlying tools constant (e.g., standard APIs) but use LLMs to synthesize a large volume of increasingly complex user queries, effectively scaling the data without changing the environment's rules or tools. Instead of static Q$\&$A, $\tau$-bench \cite{yao2024tau} employs active user simulators to test whether agents can handle dynamic conversations, while $\tau^2$-bench \cite{barres2025tau2} introduces "dual-control" mechanics where the user actively interferes with the environment state (e.g., clicking buttons while the agent speaks). Rather than sticking to one theme, AUTOENV \cite{zhang2025autoenv} automates the creation of entirely new environments with distinct rules and physics (e.g., generating 36 different games and puzzles) to test the agent's ability to adapt across heterogeneous worlds.

To our knowledge, ToolQA-D \cite{chen2024learning} is the only existing benchmark that explicitly addresses tool variability. Specifically, ToolQA-D employs GPT-4 to modify API interfaces, including names, parameters, and response formats, to evaluate robustness. 


\begin{figure*}[!t]
  \centering \centerline{\includegraphics[width=\textwidth]{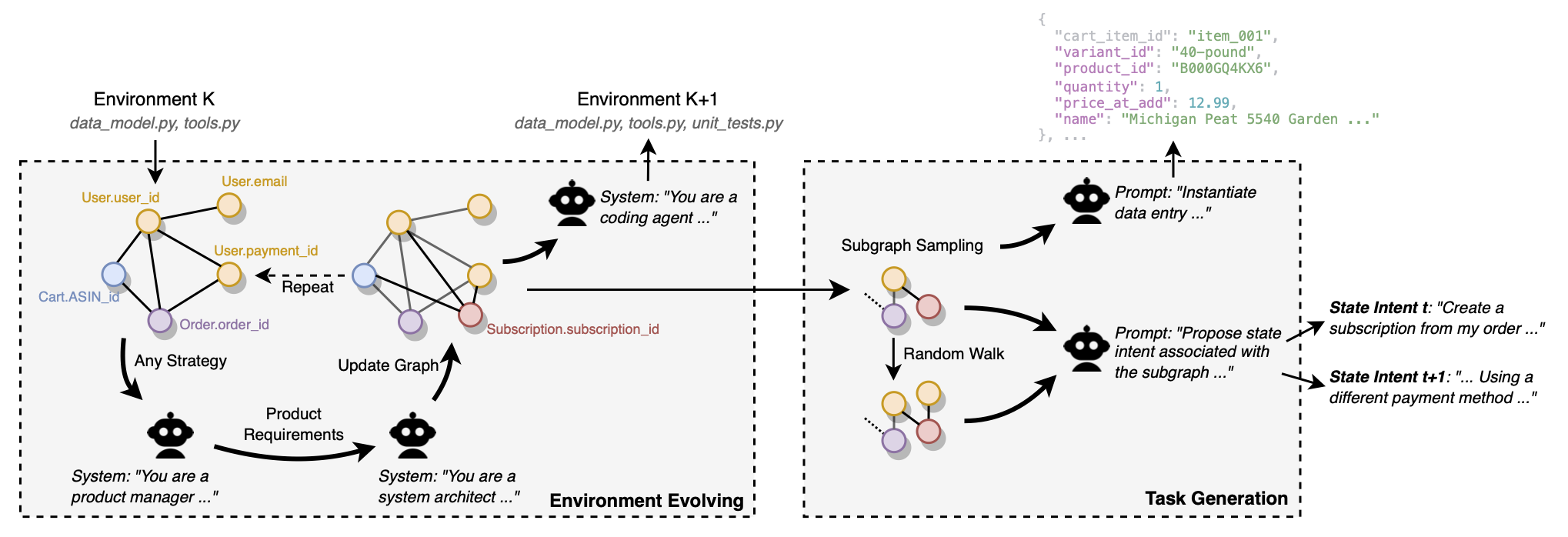}}
\vspace{-.2em}
\caption{\textbf{End-to-end workflow of \textsc{ProEvolve}.}
Starting from a seed tool-calling environment, \textsc{ProEvolve} represents schemas, data entities, and tools as an environment graph, applies semantic evolution intents as graph transformations, and instantiates the transformed graph into executable environment versions. For each version, task sandboxes are generated by sampling task subgraphs, materializing version-specific prerequisites, and producing state-wise user instructions and success criteria, enabling controlled evaluation under structured environment evolution.}

  \label{fig:evolution_workflow}
  \vspace{-1em}
\end{figure*}

\section{From Static to Evolving: Problem Setup and Challenges}

\label{sec:motivation:evolving}

Existing benchmarks typically evaluate tool-calling agents in static environments, where tools, schemas, and reachable relations remain fixed throughout evaluation. Yet real-world systems are rarely static: capabilities are introduced, interfaces are updated, and outdated functionality may be deprecated over time. Motivated by this gap, we study evolving environments, where these interfaces change across versions through explicit, dependency-aware updates. This shift from fixed snapshots to structured versioned change is the setting addressed in this section.

\subsection{Definition: Structured Environment Evolution}
\label{sec:evolution-definition}

We now formalize the notion of \emph{structured environment evolution} that
this paper studies, distinguishing it from weaker forms of environment
modification prevalent in existing benchmarks.

\begin{definition}[Structured Environment Evolution]
\label{def:evolution}
Let $\mathcal{G}^{(0)}$ be a seed environment. A \emph{structured
environment evolution} is a sequence
$\mathcal{G}^{(0)} \xrightarrow{\Delta^{(1)}} \mathcal{G}^{(1)}
\xrightarrow{\Delta^{(2)}} \cdots \xrightarrow{\Delta^{(K)}}
\mathcal{G}^{(K)}$ satisfying:
\begin{enumerate}[leftmargin=*, itemsep=2pt]
    \item \textbf{Sequential Dependence.} Each $\Delta^{(k+1)}$ is
    conditioned on current $\mathcal{G}^{(k)}$---not independently
    sampled.
    \item \textbf{Structural Memory.} $\mathcal{G}^{(k+1)}$ preserves
    all components of $\mathcal{G}^{(k)}$ not explicitly removed by
    $\Delta^{(k+1)}$, maintaining continuity across versions.
    \item \textbf{Coherent Propagation.} $\Delta^{(k+1)}$ simultaneously
    updates all coupled components (schemas, tools, data access),
    maintaining functional consistency.
\end{enumerate}

Unlike task scaling, independent environment generation, or surface-level interface perturbation, structured evolution is version-linked: each transition preserves unmodified structure and coherently propagates intended changes across the environment and its task sandboxes.

\end{definition}

This leads to the central question of the paper:
\begin{center}
\setlength{\fboxsep}{7pt}
\setlength{\fboxrule}{0.8pt}
\fbox{
\begin{minipage}{0.92\linewidth}
\emph{How to evolve environments in a controllable manner for agent benchmarking?}
\end{minipage}}
\end{center}



Answering this question requires two desiderata:

\textbf{Desideratum I: Scalable Coherence.}
Environment evolution should scale beyond manually curated updates while preserving coherent dependencies among schemas, data entities, tools, executable behavior, and task sandboxes. Naively adding tools or entities may increase benchmark size, but can break reachability, create invalid data dependencies, or produce tasks that are no longer grounded in the corresponding environment version. Thus, scalable evolution requires generating diverse environment versions without sacrificing structural and functional consistency.

\textbf{Desideratum II: Controlled Versioned Dynamics.}
Environment changes should be explicit, reproducible, and evaluable across versions. Each transition should specify what is added, reorganized, or deprecated; how the change propagates through coupled components; and how tasks remain valid under the new version. Without such control, benchmark construction reduces to unconstrained environment sampling rather than structured evolution.



These challenges motivate us to propose a programmable framework for evolving environments in a controllable and scalable manner, as described in the following sections.


\section{ProEvolve: Programmable Evolution for Agent Benchmarks}



In the following sections, we first devise a principled way to model the environment explicitly, thereby enabling programmable evolution,
as discussed in Sec.~\ref{subsec:method:graph}.
Building upon this formalism, we then propose an integrated framework to evolve environments in a programmable manner:
(i) programming environments to evolve in a coherent manner (Sec.~\ref{subsec:method:graph_evolve}),
(ii) programming sandboxes and tasks with evolved environments (Sec.~\ref{subsec:method:taskgen}),
and (iii) designing evaluation along evolution trajectories in Sec.~\ref{subsec:method:eval}. We present an overview of the framework in Figure~\ref{fig:evolution_workflow}.




\subsection{Graph Formalism for Environment Modeling}

\label{subsec:method:graph}
To operationalize structured environment evolution, we need an explicit representation of the components that co-evolve in a tool-calling environment. We therefore model schemas, data entities, tools, and their dependency relations in a unified graph representation.


To this end, we introduce a graph formalism that models environment elements and their relationships in an explicit, integrated representation. 
This choice is motivated by two merits:
1) it makes reachable information explicit by encoding schemas, data entities, tools, and their relations in a shared structure; 2) it makes version-linked evolution programmable: graph transformations can add, reorganize, or remove capabilities while preserving unmodified structure and propagating changes through coupled components.

 
Concretely, we represent an environment version through a typed relational graph
$\mathcal{G}=(\mathcal{V},\mathcal{E})$.
Each node $v\in\mathcal{V}$ corresponds to a schema element (e.g., \texttt{User.user\_id}, \texttt{Order.order\_id}),
and each directed edge $e=(v\!\rightarrow\!v')\in\mathcal{E}$ denotes a typed relation or a tool-enabled transition that maps information from a source schema element to a target schema element.
\begin{wrapfigure}{R}{0.5\textwidth}
  \vspace{-10pt}
  \centering
\includegraphics[width=\linewidth,height=0.26\textheight,keepaspectratio]{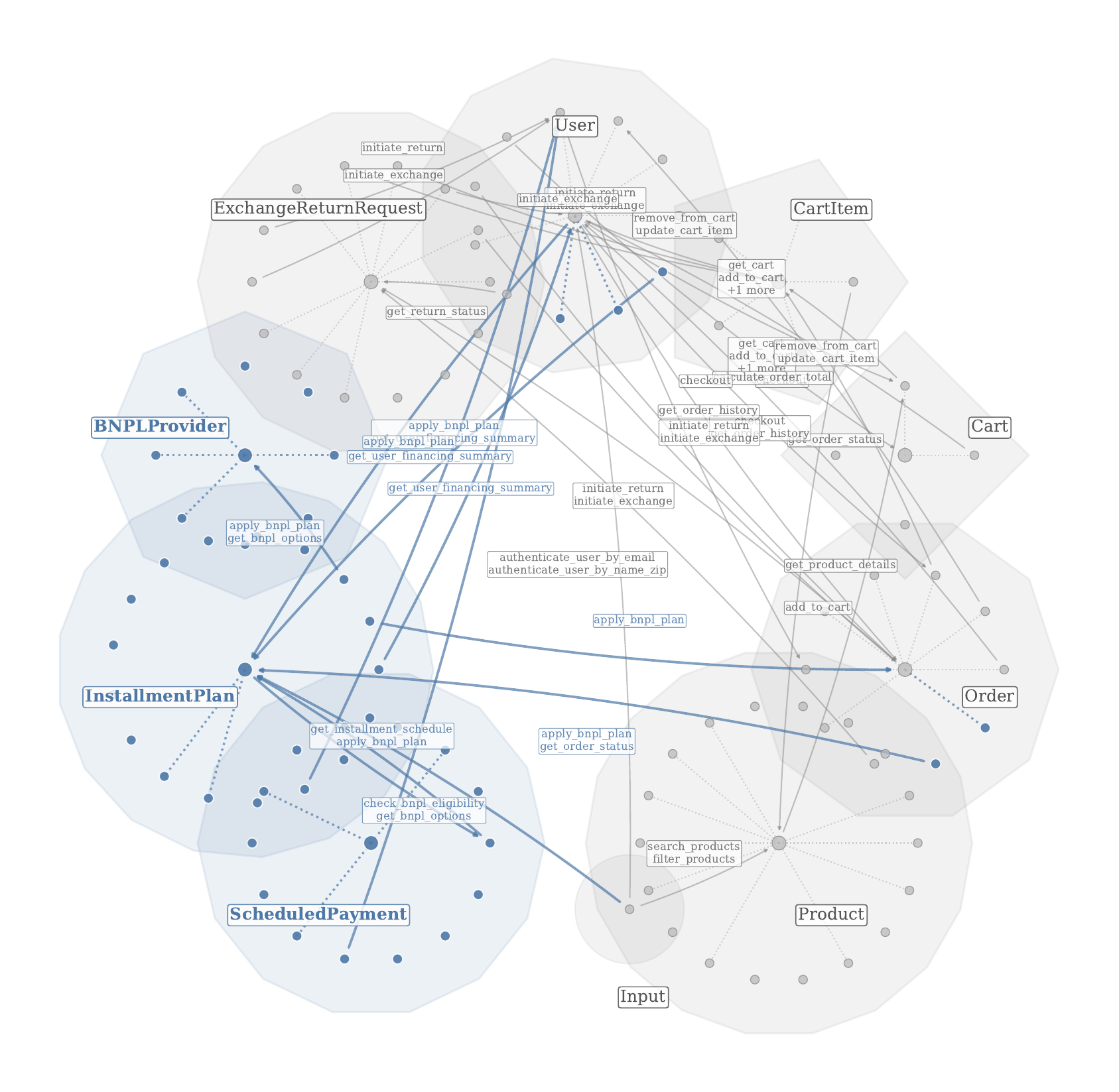}
\vspace{-18pt}
  \caption{\textbf{Programmed Environment Evolution with two versions} The evolution is programmed via graph transformations, expressed as editions on nodes (schemas, e.g., \texttt{InstallmentPlan}) and edges (tool-enabled transitions, e.g., \texttt{get\_installment\_schedule}) that derives evolutions with existing structure. 
  }
  \label{fig:graph_evolution}
  \vspace{-10pt}
\end{wrapfigure}

As such, environment evolution becomes a sequence of graph transformations:
\begin{equation}
\mathcal{G}^{(0)} \xrightarrow{\;\Delta^{(1)}\;} \mathcal{G}^{(1)} \xrightarrow{\;\Delta^{(2)}\;} \cdots \xrightarrow{\;\Delta^{(K)}\;} \mathcal{G}^{(K)},
\label{eq:graph_evolution_seq}
\end{equation}
where $\mathcal{G}^{(0)}$ is a seed environment and each $\Delta^{(k)}$ applies a structured operation to the current version. Each transformation is version-linked: it is conditioned on $G^{(k-1)}$, preserves components not explicitly modified, and updates affected relations so the resulting environment remains executable and evaluable.
Fig.~\ref{fig:graph_evolution} illustrates this process. Adding a new capability is not treated as an isolated tool insertion: the graph edit introduces the required entities and attributes, connects them to existing schemas via dependency edges, and exposes the corresponding tool-enabled transitions. This makes the intended change explicit before it is implemented as code.


The following sections instantiate this formalism along three steps: programming graph transformations to produce executable environment versions (Sec.~4.2), generating version-linked task sandboxes from each evolved graph (Sec.~4.3), and evaluating agents through graph-grounded, information-centric criteria (Sec.~4.4).



\subsection{Programming Environment Evolution via Graphs} 
\label{subsec:method:graph_evolve}

Whereas the graph formalism enables \emph{controllable} evolution through graph transformations (Eq.~\ref{eq:graph_evolution_seq}),
manually composing coherent graph transformations $\Delta^{(k)}$ for diverse evolution patterns does not scale. To enable scalable and diverse environment generation, we introduce an agentic pipeline that \textit{automatically} programs evolutions in two phases:

\textbf{Phase I: Evolution Proposal}: An LLM agent will traverse the environment graph to make a reasonable transformation plan based on predefined transformation strategies. 
\begin{itemize}[leftmargin=1.2em,itemsep=1pt,topsep=2pt]



    \item \texttt{Completion} ($\Delta^{\text{comp}}$) introduces new capabilities by adding the required schema elements, data entities, and tool-enabled transitions. For example, adding buy-now-pay-later support requires new financing entities, eligibility relations, and checkout-related tools.
    
    \item \texttt{Saturation} ($\Delta^{\text{sat}}$) reorganizes access paths by adding shortcut tool-enabled transitions over useful multi-hop relations. For example, a multi-step path from user identity to purchased products can be compressed into a direct purchase-history tool.
    
    \item \texttt{Deprecation} ($\Delta^{\text{dep}}$) removes or disables nodes and edges to model controlled capability removal or access loss, while preserving valid alternative paths when required.

\end{itemize}


These primitives are not intended to exhaustively cover all possible environment changes. Instead, they instantiate common structured changes in tool-calling environments: capability addition, access-path reorganization, and capability removal. Additional primitives can be incorporated as long as they can be expressed as semantic intents and graph operations under the same validation pipeline.

\textbf{Phase II: Implement and Validate}:

Given the transformed graph $\mathcal{G}^{(k)}$, ProEvolve instantiates an executable environment version by updating schemas, data models, tool implementations, and graph-grounded tests. The graph serves as the structural specification for code generation, ensuring that implementation changes are derived from explicit dependencies rather than disconnected natural-language edits.

Each generated version is validated along three axes. \textbf{Graph validity} checks whether the graph edits preserve schema consistency, reachability, and dependency constraints. \textbf{Implementation validity} checks whether the generated code is executable and passes graph-grounded unit tests. \textbf{Semantic-intent alignment} checks whether the implemented change realizes the intended evolution $\Delta^{(k)}$, rather than merely passing tests. Versions that fail critical checks are filtered before being used for task generation or downstream evaluation.


This graph-first design addresses the two desiderata in Section~3. First, representing each evolution step as explicit graph edits preserves scalable coherence: new capabilities, schema changes, and tool-enabled transitions are introduced through a unified dependency structure rather than as disconnected modifications. Second, graph-grounded specifications support quality assurance before a version is advanced to the next stage: the pipeline checks executability, structural consistency, and semantic alignment with the intended evolution. This step-wise validation helps contain error accumulation across successive environment versions.

\begin{wrapfigure}{R}{0.44\textwidth}
  \vspace{-10pt}
  \centering
  \includegraphics[width=\linewidth,height=0.26\textheight,keepaspectratio]{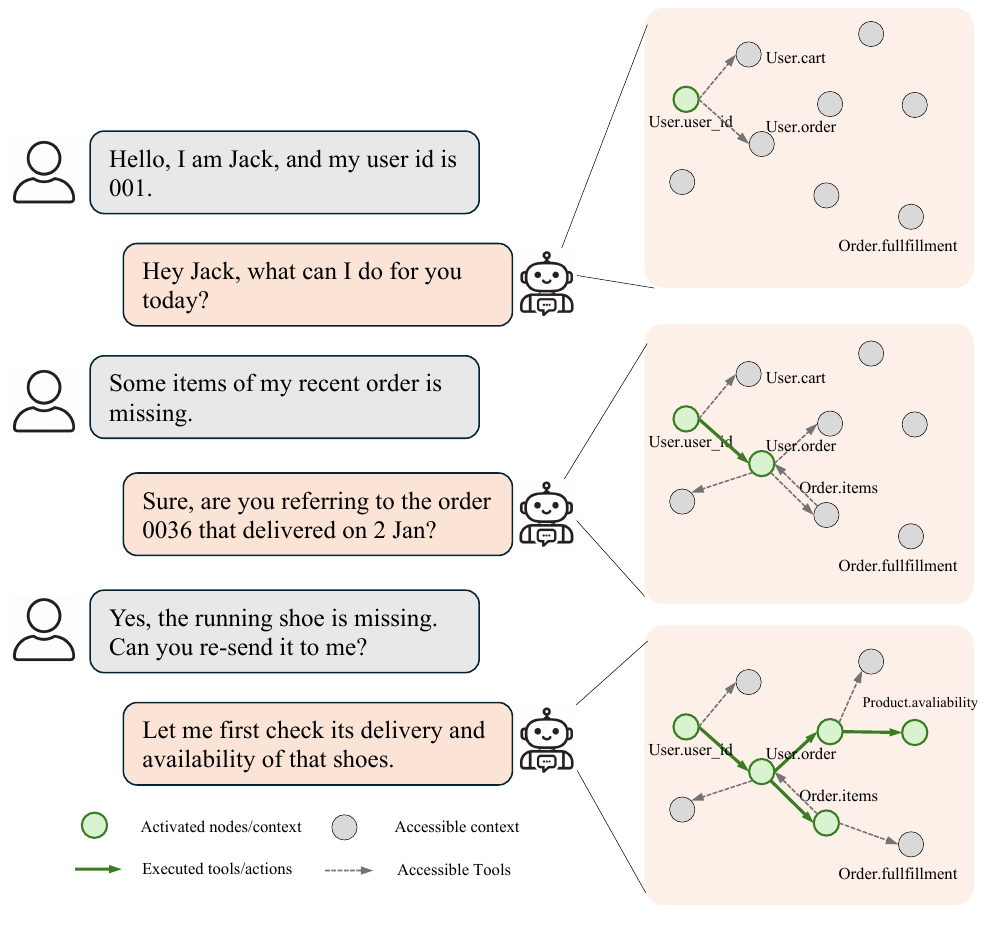}
  \caption{\textbf{Context as subgraph expansion in a tool-mediated conversation.}
At each turn, the environment exposes a reachable context subgraph (gray nodes; dashed arrows for reachable tool transitions),
while the agent activates a subset of nodes (green) by executing tools/actions (solid arrows) conditioned on the dialogue.
As the conversation progresses, executed transitions expand the active subgraph, enabling retrieval and integration of newly reachable
information (e.g., from \texttt{User.user\_id} to \texttt{User.order} to \texttt{Order.order\_items} and downstream product attributes).}
  \label{fig:graph_expansion}
  \vspace{-25pt}
\end{wrapfigure}

\subsection{Programming Tasks as Subgraphs}
\label{subsec:method:taskgen}

Given an evolved environment version $\mathcal{G}^{(k)}$, task generation must remain grounded in the capabilities and constraints of that version. We therefore formulate each task as a constrained subgraph $\mathcal{G}_{\tau} \subseteq \mathcal{G}^{(k)}$, which specifies the schemas, data dependencies, and tool-enabled transitions needed to satisfy the task objective. Since $\mathcal{G}_{\tau}$ is sampled from the current graph, each task reflects the tools, reachable information, and constraints available in that version, making task sandboxes version-linked rather than drawn from a environment-agnostic generation. 




The task programming consists of the following steps: 

\begin{itemize}[leftmargin=1.2em,itemsep=1pt,topsep=2pt]
    \item \textbf{\texttt{Subgraph Sampling.}}
    We sample a connected task subgraph  $\mathcal{G}_{\tau} \subseteq \mathcal{G}^{(k)}$ and synthesize a task-level goal $g_\tau$ and scenario description $s_\tau$ conditioned on the sampled structure.


    \item \textbf{\texttt{Sandbox Materialization.}}
Given $(\mathcal{G}_{\tau}, s_{\tau})$, we instantiate prerequisite entities, attributes, and cross-references according to the relations in $\mathcal{G}_\tau$, ensuring that required tool transitions are executable under the current environment version. When a scenario involves multiple prerequisites, we compose them by jointly instantiating the prerequisite nodes and edges induced by $\mathcal{G}_\tau$ and enforcing shared identifiers and cross-entity references.

    \item \textbf{\texttt{Agentic Walk Execution.}}
    We then perform an oracle traversal over $\mathcal{G}_\tau$ to produce state-wise user instructions, graph-derived success criteria, and a reference trace of tool actions and activated graph elements. Tool outputs are converted into newly obtained facts and accumulated as:
    \begin{equation}
    \widehat{\mathcal{K}}_{t} \;=\; \widehat{\mathcal{K}}_{t-1} \cup \Delta\widehat{\mathcal{K}}_{t},
    \qquad
    \Delta\widehat{\mathcal{K}}_{t} \;=\; \phi(\mathbf{y}_t).
    \label{eq:kt_update}
    \end{equation}
    Meanwhile, the active task context is represented as a subgraph $\mathcal{A}_t \subseteq \mathcal{G}_{\tau}$ that expands with executed actions and revealed information:
    \begin{equation}
    \mathcal{A}_{t} \;=\; \mathrm{Expand}\!\Big(\mathcal{A}_{t-1};\ a_t,\ u_t^\star,\ \mathcal{G}_{\tau}\Big),
    \label{eq:At_expand_taskgen}
    \end{equation}
    yielding a sequence of structured expansions
    $\mathcal{A}_0 \rightarrow \mathcal{A}_1 \rightarrow \cdots \rightarrow \mathcal{A}_T$.
\end{itemize}



\paragraph{Sandbox output.}
Each task instance $\tau$ produces a version-specific sandbox containing: (1) an initial state $\mathcal{S}_0$ with materialized prerequisite entities and cross-references; (2) state-wise user instructions $\{u_t^\star\}_{t=1}^T$ paired with graph-derived success criteria $\{\mathcal{Y}_t^\star\}_{t=1}^T$; and (3) a reference graph trajectory recording oracle actions, activated nodes and edges, and context expansions. Since all components are derived from $\mathcal{G}_\tau \subseteq \mathcal{G}^{(k)}$, each sandbox remains grounded in the capabilities, constraints, and reachable information of its corresponding environment version.

\subsection{State-Wise User Simulation and Evaluation}
\label{subsec:method:eval}

Under structured evolution, tool names, schemas, and access paths may change across versions, so exact tool-sequence matching becomes brittle. We therefore evaluate task progress at the level of graph-derived information: whether the agent obtains and uses the facts required by each state of the task.

\textbf{State instruction and criterion.}
For each state $t$, we construct an instruction tuple
\begin{equation}
\mathcal{I}_t \;=\; \Big(u_t^\star,\; \mathcal{Y}_t^\star\Big),
\label{eq:state_instruction}
\end{equation}
where $u_t^\star$ is the synthesized user instruction and $\mathcal{Y}_t^\star$ denotes the minimal set of graph-grounded facts required to satisfy the state. In practice, $\mathcal{Y}_t^\star$ is derived from the frontier targets and extracted facts in the reference trajectory, so the criterion is grounded in the task subgraph $\mathcal{G}_\tau \subseteq \mathcal{G}^{(k)}$ .


\textbf{Progression rule.}
The user simulator follows the state-wise instructions and advances from state $t$ to $t+1$ only if the agent satisfies $\mathcal{Y}_t^\star$, judged by schema-aware extraction from tool outputs and response text. Otherwise, the simulator remains at the current state and issues a follow-up clarification.


\textbf{State success rate.}
Let $s_t \in \{0,1\}$ denote if the agent satisfies state $t$. The state-wise success rate is:
\[
\mathcal{C}(\tau)=\frac{1}{T}\sum_{t=1}^{T}s_t.
\]
Because each $Y_t^\star$ and instruction $u_t^\star$ are derived from the task subgraph expansion (Eq.~\ref{eq:At_expand_taskgen}), $\mathcal{C}(\tau)$ measures dependency-consistent progress through the evolved environment. This yields stable user simulation and supports an information-centric, tool-agnostic evaluation: rather than matching fixed tool sequences or explicit tool annotations, we assess whether the agent acquires and uses the required information. This is especially suitable for structured evolution, where tools, schemas, or access paths may change while equivalent task progress remains valid.

\section{Experiments}
\label{sec:experiments}

We evaluate ProEvolve primarily as a benchmark-construction framework. Our experiments are organized around three questions:





\textbf{Q1. Validity:} Are evolved environments executable and semantically aligned?

\textbf{Q2. Quality:} Are environments and task sandboxes diverse, version-linked, and nontrivial?

\textbf{Q3. Utility:} Do evolved environments induce meaningful agent-behavior differences?




\subsection{Benchmark Construction Setup}
\label{subsec:bench_setup}

\leavevmode
\begin{wraptable}{r}{0.48\textwidth}
\renewcommand{\arraystretch}{0.94}
\vspace{-20pt}
  \centering
  \small
  \caption{ Scale/Quality of evolved environments.}
  \label{tab:benchmark-detail}
  \begin{tabular}{lrr}
    \toprule
    & \textbf{E-com.} & \textbf{Airline} \\
    \midrule
    \multicolumn{3}{l}{\textit{Benchmark Scale}} \\
    \quad Seed environments       & 1     & 1     \\
    \quad Evolution trajectories  & 50    & 5     \\
    \quad Environment variants    & 200   & 20    \\
    \quad Task sandboxes          & 3{,}000 & 300 \\
    \midrule
    \multicolumn{3}{l}{\textit{Environment Complexity}} \\
    \quad Unique tools            & 384   & 58    \\
    \quad Unique schemas          & 201   & 22    \\
    \quad Seed tools / schemas    & 51/64 & 21/8 \\
    \midrule
    \multicolumn{3}{l}{\textit{Implementation Validity}} \\
    \quad Unit test coverage      & 100\% & 100\% \\
    \quad Unit test pass rate     & 90.8\%& 93.1\%\\
    \quad Defect-free env rate     & 96.0\%& 86.7\%\\
    \midrule
    \multicolumn{3}{l}{\textit{Semantic Fidelity\;{\scriptsize(1--5)}}} \\
    \quad Reality                 & 4.76  & 4.78  \\
    \quad Graph fidelity          & 4.63  & 4.71  \\
    \quad Code fidelity           & 4.55  & 4.60  \\
    \midrule
        \multicolumn{3}{l}{\textit{Sandbox Quality\;{\scriptsize(1--5)}}} \\
    \quad Scenario Coherence	 &4.75	 &4.81\\ 
    \quad Difficulty Calibration	 & 5.00	 & 5.00 \\
    \quad State Transition Feasibility & 	4.50 & 	4.06 \\
    \quad Prerequisite Sufficiency	 & 4.62	 & 4.88 \\
    \midrule
        \multicolumn{3}{l}{\textit{Environment Diversity}} \\
    \quad Pairwise tool Jaccard\,$\uparrow$  & \multicolumn{2}{c}{0.976} \\
    \quad Unique tool ratio\,$\uparrow$      & \multicolumn{2}{c}{0.645} \\
    \quad Tool freq.\ entropy\,$\uparrow$    & \multicolumn{2}{c}{8.07}  \\
    \bottomrule
    
  \end{tabular}
  \vspace{-30pt}
\end{wraptable}

\textbf{Domains.}
We construct benchmarks in two structured tool-calling domains. E-commerce serves as our main large-scale validation domain: it contains 1,000 products sourced from WebShop~\cite{yao2022webshop}, 50 synthesized users, 51 seed tools, and 64 seed schemas. Airline booking serves as a cross-domain pilot adapted from $\tau^2$-bench~\cite{barres2025tau2} to test whether the same construction pipeline generalizes beyond e-commerce. For each seed environment, we construct an environment graph from the available schemas, data entities, and tools.

\textbf{Environment Evolution.}
Starting from each seed graph, we instantiate evolution trajectories using the primitive sequence, i.e.,
$\mathcal{G}^{(0)}
  \xrightarrow{\Delta^{\text{comp}}} \mathcal{G}^{(1)}
  \xrightarrow{\Delta^{\text{sat}}}  \mathcal{G}^{(2)}
  \xrightarrow{\Delta^{\text{dep}}}  \mathcal{G}^{(3)}$
corresponding to capability addition, access-path reorganization, and controlled capability removal. 
We generate 50 e-commerce trajectories and 5 airline trajectories, each containing four environment versions, yielding 200 and 20 version-linked environments respectively.

\textbf{Task Generation.} 
For each environment version, we generate task sandboxes using the subgraph-based pipeline in Sec.~\ref{subsec:method:taskgen}. Each task is sampled from the corresponding evolved graph and materialized with version-specific entities, prerequisites, state-wise instructions, and graph-derived success criteria. We balance task difficulty with an equal ratio of easy, medium, and hard tasks.


\textbf{Downstream agent evaluation setup.}
For downstream diagnostic evaluation, we run representative tool-calling agents using the same user-simulation protocol described in Sec.~\ref{subsec:method:eval}. We report state-wise success rate $C$, average turns $\bar{T}$, and average tool calls $\bar{N}_{\mathrm{tool}}$. 

\subsection{Generation Quality of Evolved Environments and tasks}
\label{subsec:gen_quality}

Before benchmarking agents, we assess the generated environments and task sandboxes along four axes: implementation validity, semantic-intent alignment, environment diversity, and task difficulty calibration. Summary statistics are reported in Table~\ref{tab:benchmark-detail}; detailed breakdowns are in Appendix~\ref{app:gen-quality}.

\textbf{Implementation Validity.}
\label{subsubsec:impl_validity}
Across all 200 e-commerce environments, unit tests achieve 100\%
line coverage over modified code with a 90.8\% pass rate; the
airline domain yields 93.1\%. More informatively,
{96.0\%} of e-commerce env-versions and {86.7\%} of airline
env-versions are \emph{defect-free}, containing zero failures that
indicate actual tool-code bugs. A systematic analysis of the remaining
failing cases reveals that the majority are benign: 51\% stem from test
implementation bugs (not environment code), 23\% are corner-case checks
with minimal usage impact, and only 10\% (${\sim}1\%$ of all tests)
affect actual tool behavior. Environments flagged with tool defects are
filtered before the next evolution step, preventing error accumulation.
Full categorization is in Appendix~\ref{app:test-failure}.

\begin{table}[t]

\centering\scriptsize
\caption{Baseline behavior across e-commerce versions. Each cell reports $\mathcal{C}(\Delta)/\bar{T}/\bar{N}_{\mathrm{tool}}$
(state-wise success / avg turns / avg tool calls).}\label{tab:versionwise-baseline}
\begin{tabular}{lcccc}
\toprule
Model & Seed & $\Delta^{\text{comp}}$ & $\Delta^{\text{sat}}$ & $\Delta^{\text{dep}}$ \\
\midrule
GPT-5  ~\cite{singh2025openai}    & 0.56 / 8.1 / 8.1  & 0.65 \pos{0.08} / 10.3 / 12.8 & 0.77 \pos{0.21} / 13.2 / 14.9 & 0.45 \nega{0.11} / 8.7 / 20.6 \\
Claude-Opus-4.5 ~\cite{claude}   & 0.49 / 6.3 / 4.7  & 0.46 \nega{0.02} / 5.5 / 4.8  & 0.62 \pos{0.14} / 7.3 / 5.9   & 0.37 \nega{0.12} / 4.5 / 4.4  \\
DeepSeek-v3.2 ~\cite{liu2025deepseek}   & 0.53 / 7.3 / 6.7  & 0.52 \nega{0.01} / 7.4 / 6.9  & 0.49 \nega{0.04} / 8.5 / 6.7  & 0.46 \nega{0.07} / 9.0 / 7.3  \\
Qwen3-235B  ~\cite{qwen3}   & 0.59 / 10.4 / 9.5 & 0.63 \pos{0.05} / 12.9 / 13.1 & 0.46 \nega{0.13} / 9.0 / 8.1  & 0.36 \nega{0.22} / 7.6 / 7.4  \\
Gemini-2.5-Pro ~\cite{comanici2025gemini}    & 0.43 / 7.3 / 3.6  & 0.57 \pos{0.15} / 8.9 / 5.1   & 0.54 \pos{0.12} / 8.0 / 3.9   & 0.45 \pos{0.03} / 8.1 / 3.9  \\
\bottomrule
\end{tabular}
\vspace{-12pt}
\label{tab:baseline}
\end{table}



\textbf{Semantic Fidelity.}
\label{subsubsec:semantic_fidelity}
We verify that each evolution step faithfully realizes its intended change
using an LLM-as-a-Judge applied to each individual modification.
(plausibility as a real-world change), \textit{Graph Fidelity} (alignment
between implemented code and specified graph edits), and \textit{Code
Fidelity} (correctness of tool signatures and cross-entity relations).
No significant degradation is observed across all three evolution strategies. Rubric definitions and per-strategy distributions are in
Appendix~\ref{app:semantic-fidelity}.

\begin{figure}[!t]
  \centering
  \includegraphics[width=.87\linewidth,height=0.4\textheight,keepaspectratio]{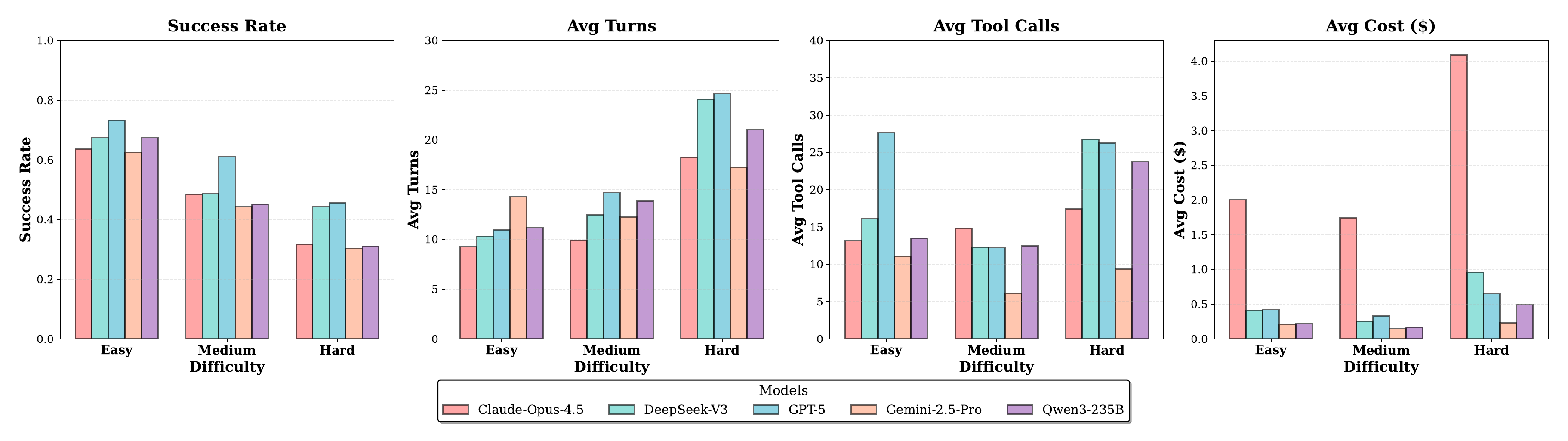}
  \vspace{-7pt}
  \caption{\textbf{Efficiency breakdown by task difficulty.}
We report average tool calls, estimated cost, conversation turns, and reward for each model under varying difficulties.
}
\label{fig:metrics_by_difficulty}
\vspace{-8pt}
\end{figure}

\textbf{Environment Diversity.}
\label{subsubsec:env_diversity}
From a seed with 51 tools and 64 schemas, the 200 e-commerce environments
span {384 unique tools} and {201 unique schemas}, with a mean
pairwise tool Jaccard distance of {0.976} (near-zero overlap across
environments). Compared to a non-graph baseline that evolves from the
same seed without graph-guided transformations, \textsc{ProEvolve} yields
higher diversity across all metrics ($+0.109$ Jaccard distance, $+0.251$
unique tool ratio, $+0.41$ bits entropy). This diversity is further
justified by the substantial environment-to-environment variability in agent performance reported in Sec.~\ref{subsec:env_evolution}. Full
comparison is in Appendix~\ref{app:diversity}.


\textbf{Sandbox Quality.}
We verify that graph-grounded difficulty levels produce meaningful behavioral separation. As shown in Figure~\ref{fig:metrics_by_difficulty}, success decreases as difficulty increases, while turns and tool calls generally grow, reflecting deeper graph traversal; for example, \texttt{GPT-5} drops from ${\sim}0.80$ on easy tasks to ${\sim}0.62$ on hard tasks with nearly doubled tool usage, while \texttt{Gemini-2.5-Pro} remains cost-frugal but performs worse on hard tasks, suggesting under-exploration. 

We further assess the intrinsic quality of generated task sandboxes using an LLM-as-a-Judge over four dimensions. Across 200 sampled sandboxes balanced by difficulty and evolution primitive, the mean scores are 4.75, 5.00, 4.50, and 4.62, with 0\% scoring $\leq 2$ on any dimension. These results indicate that subgraph complexity provides effective difficulty control and that generated tasks are coherent, well-calibrated, and executable under evolving environments.

\subsection{Agent Evaluation as Downstream Diagnostic}
\label{subsec:env_evolution}


After validating generation quality, we use agent evaluation as a
downstream diagnostic. The evolved environments enable two analyses
that are difficult to conduct with independently sampled benchmarks:
(1)~\emph{counterfactual capability measurement} via version-linked
comparisons, and (2)~\emph{memory-based adaptation probes} under
structured change. Full results are present in Appendix~\ref{app:agent-results}.

\textbf{Counterfactual capability measurement.}
Because each environment version is derived from the same seed via a
single, controlled graph edit, comparing agent performance across
versions yields counterfactual estimates of specific capabilities, which cannot be done with two independently generated environments.
Table~\ref{tab:baseline} reports performance changes across each evolution trajectory:

\begin{itemize}[nosep,leftmargin=*]
\item \textbf{Scalability}
  (Seed\,$\to$\,$\Delta^{\text{comp}}$): measures how agents scale
  to richer environments. Most models maintain or improve
  $\mathcal{C}$ (GPT-5: $+0.08$, Gemini-2.5-Pro: $+0.15$), indicating that
  reasonable complexity increases remain tractable, but at the cost
  of added traversal depth (higher $\bar{T}$ and $\bar{N}$).
\item \textbf{Efficiency}
  (Seed\,$\to$\,$\Delta^{\text{sat}}$): measures how agents exploit
  shortcut access paths. Most models benefit from the expanded action
  space, achieving higher completion (GPT-5: $+0.21$, Claude-Opus-4.5:
  $+0.14$). However, the expanded option set does not uniformly improve efficiency across models
  turns and tool calls increase across the board, as agents explore
  among multiple valid paths rather than converging on the shortest
  one.
\item \textbf{Robustness}
  (Seed\,$\to$\,$\Delta^{\text{dep}}$): examines the recovery when tools
  are deprecated. All models except Gemini-2.5-Pro declines in $\mathcal{C}$,
  with Qwen3-235B showing the largest drop ($-0.22$). Notably, GPT-5's
  tool calls spike  from 8.1 to 20.6 despite lower success,
  revealing unrecovered retry loops on deprecated tools.
\end{itemize}

\begin{wrapfigure}{rt}{0.38\textwidth}
  \vspace{-10pt}
  \centering
\includegraphics[width=\linewidth,height=0.32\textheight,keepaspectratio]{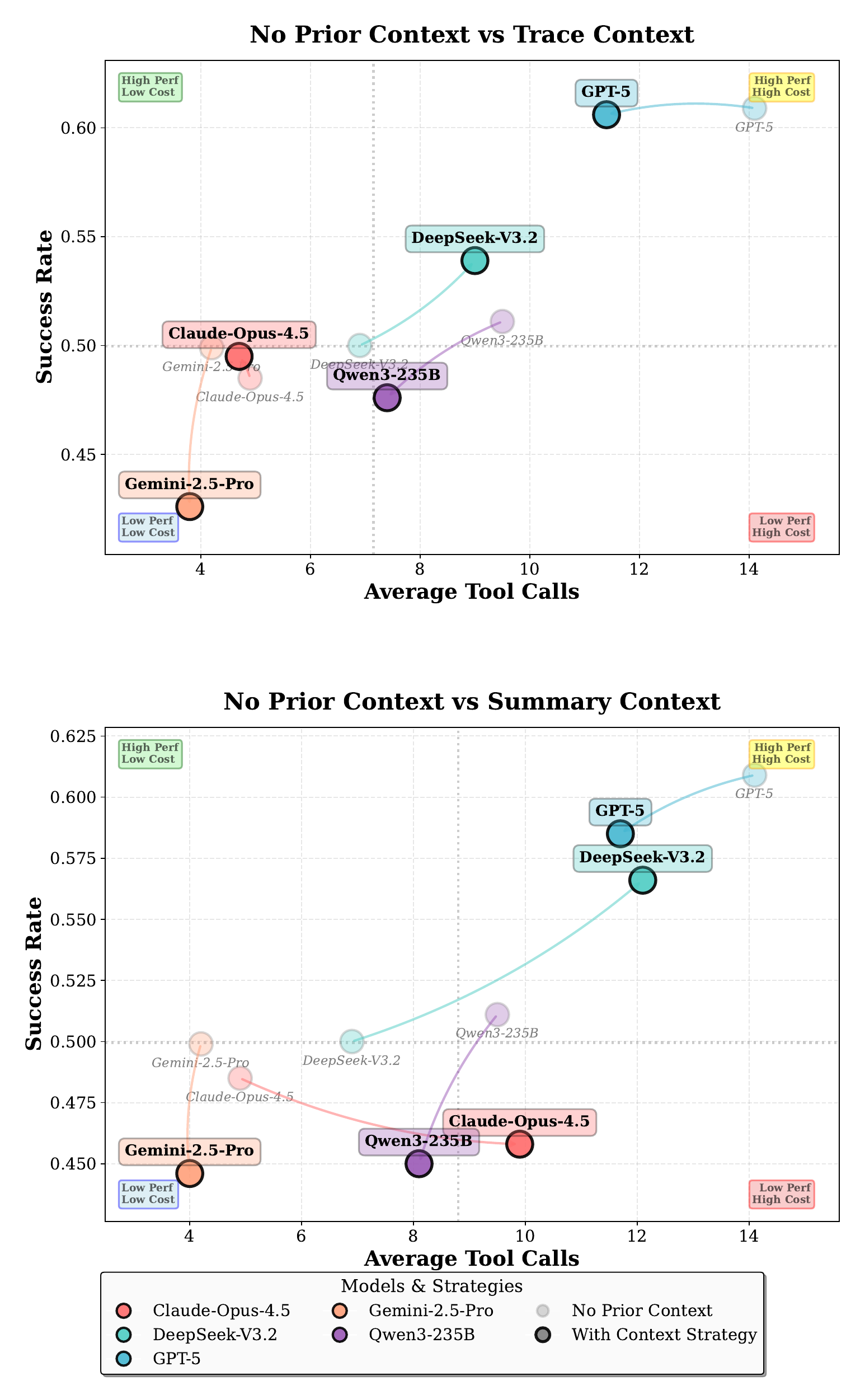}
\caption{Performance--efficiency tradeoff under context priors. Each point represents an agent/probe setting, with state-wise success on the y-axis and average tool calls on the x-axis. 
}
\label{fig:cost_perf_tradeoff}
  \vspace{0pt}
\end{wrapfigure}


\textbf{Memory-based adaptation.}
Memory-augmented agents are typically developed and evaluated in
static environments. We probe whether two common memory strategies,
\emph{Trace Context} (raw interaction history) and
\emph{Summary Context} (distilled natural-language summaries),
transfer to evolving settings
(Figure~\ref{fig:cost_perf_tradeoff}). Two patterns emerge. First,
\textbf{\textit{memory improves efficiency but often hurts task
completion}}: most models reduce tool calls under context replay
(arrows move left) but do not improve, or actively degrade, on
success rate, indicating that prior-version experience becomes stale
or misleading after structural changes. The exception is
DeepSeek-V3.2, which benefits from both context variants at the cost
of higher tool usage. Second, \textbf{\textit{Trace and Summary
Context yield similar trends for most models}}, with one notable
exception: Claude-Opus-4.5 improves under Trace Context (higher
success, comparable cost) but degrades under Summary Context (more
tool calls, lower success), suggesting that distilled summaries
over-generalize deprecated tool availability and induce
over-exploration.

\begin{wrapfigure}{r}{0.36\textwidth}
  \centering
   \vspace{-20pt}
\includegraphics[width=\linewidth,height=0.4\textheight,keepaspectratio]{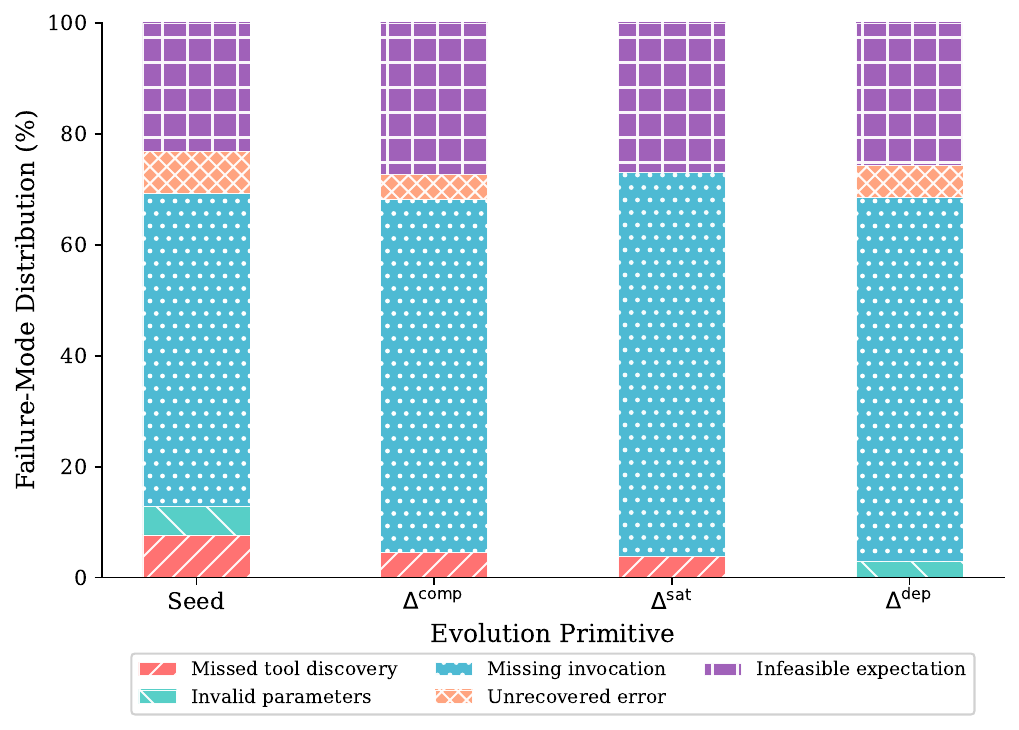}
  \caption{
  Failure-mode breakdown across environment versions. Bars decompose sampled failed trajectories by coarse error type.
  }
  \label{fig:failure_modes}
\end{wrapfigure}

\textbf{Failure modes align with graph transformations.}
Figure~\ref{fig:failure_modes} decomposes agent failures by environment version. Across all versions, wrong or obsolete tool use is the dominant failure mode, suggesting that agents primarily struggle to identify valid access paths as tools, schemas, and reachable relations change. Unrecovered tool errors and retry loops indicate weaker recovery behavior after failed calls, while task/prerequisite misunderstanding reflects failures in tracking graph-grounded task dependencies. The versioned breakdown shows that failures are interpretable with respect to the evolving environment structure, supporting the diagnostic utility of \textsc{ProEvolve}.

\section{Conclusions}
\label{sec:conclusions}

We study structured environment evolution as a benchmark-construction problem for tool-calling agents, formalizing it as version-linked, dependency-aware changes across schemas, tools, and task sandboxes. We introduce \textsc{ProEvolve}, a graph-based framework that programs such changes through explicit graph transformations and version-linked task generation.

Experiments in e-commerce and airline booking show that \textsc{ProEvolve} generates executable, semantically aligned, and diverse evolved environments, scaling to 200 e-commerce versions and 3,000 task sandboxes. Downstream evaluation shows that programmable factors such as evolution direction, task difficulty, and evolution pattern already expose distinct agent performance, efficiency, and failure patterns. We plan to extend this factor space to policy changes, permission constraints, tool-behavior drift, and user-behavior shifts, supporting more comprehensive evaluation under structured evolution.

\clearpage

\bibliography{example_paper}

\clearpage 

\appendix

\renewcommand\thesection{\Alph{section}}
\label{app:toc-v4}
\medskip


\noindent\textbf{Appendix Table of Contents}
\vspace{0.5em}
\noindent
\begin{tabbing}
  Appendix A \quad \= Generation Quality Analysis \kill
  Appendix \ref{app:gen-quality} > Generation Quality Analysis \\
  Appendix \ref{app:agent-results} > Full Agent Evaluation Results \\
  Appendix \ref{app:impact} > Broader Impact \\
  Appendix \ref{app:task-generation} > Version-Linked Task Sandbox Generation \\ 
  Appendix \ref{app:license} >Licenses for Existing Assets \\
  Appendix \ref{app:evolution-examples} > Illustrative Examples of Structured Environment Evolution \\
  Appendix \ref{app:prompts-implementation} > Prompts and Implementation Details \\
\end{tabbing}

\bigskip

\begin{figure*}[h]
  \centering
  \centerline{\includegraphics[width=\textwidth]{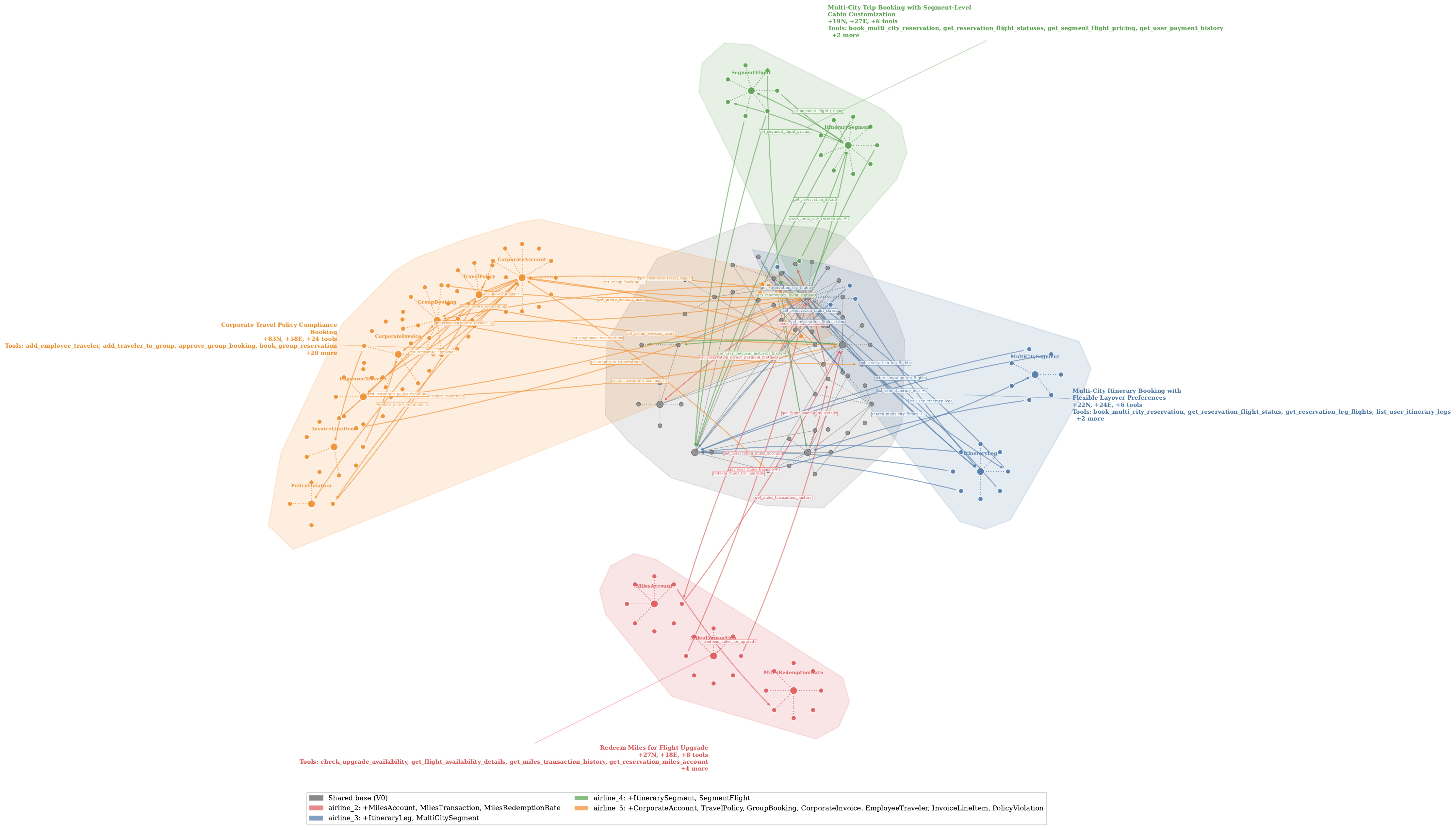}}
  \caption{Layered topology of evolved airline environments. Grey nodes and edges represent the shared seed environment (V0: 10 databases, 56 nodes, 68 edges, 21 tools). Each colored region represents an independent evolution trajectory extending the base airline schema with new capabilities. Dotted edges denote intra-database connections; solid directed edges denote cross-database tool invocations. Node labels indicate newly added databases; edge labels show tool names introduced by each trajectory.}
\label{fig:topology_airline}
\vspace{-1pt}
\end{figure*}

\begin{figure*}[h]
  \centering
  \centerline{\includegraphics[width=\textwidth]{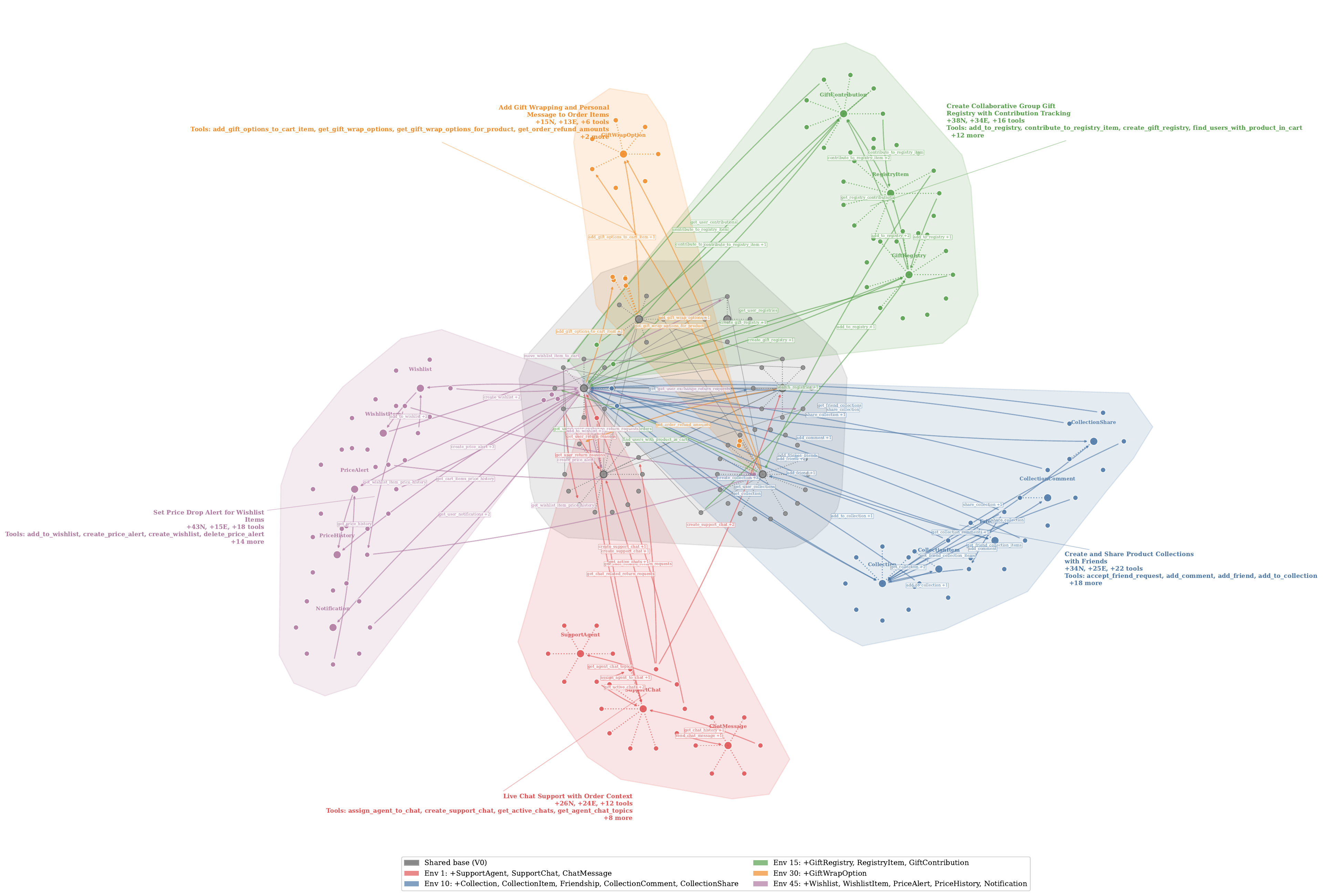}}
  \caption{Layered topology of evolved e-commerce environments. Grey nodes and edges represent the shared seed environment (V0: 7 databases, 64 nodes, 63 edges, 23 tools). Each colored region corresponds to a distinct evolution trajectory that adds new databases, attributes, and tool-mediated relationships to the base schema. Dotted edges denote intra-database attribute connections; solid directed edges denote cross-database tool invocations. Annotations on nodes indicate newly introduced databases; annotations on edges show the corresponding tool names.}
\label{fig:topology_ecomm}
\vspace{-1pt}
\end{figure*}

\section{Generation Quality Analysis}
\label{app:gen-quality}

\subsection{Implementation Validity and Test-Failure Breakdown}
\label{app:test-failure}
We categorize every failing unit test by the severity of its underlying cause, reflecting whether the failure indicates a genuine defect in the tool implementation or only a test-side artifact. The taxonomy has three levels:

\begin{itemize}
  \item \textbf{(1) No impact.} The tool implementation is correct; the test itself carries a bug. Typical signals include string-label or enum casing mismatches, hardcoded entity identifiers absent from the fixture, pydantic validation errors in fixture setup, and attribute access on \texttt{MagicMock} / \texttt{FixtureFunctionDefinition} objects.
  \item \textbf{(2) Corner case.} The tool works on common inputs but fails on an edge case of limited practical impact---an optional field not populated, a \texttt{ValueError} on an uncommon argument combination, or an assertion on a seldom-used return shape.
  \item \textbf{(3) Partial impact (tool defect).} The tool implementation is genuinely broken. Typical signals are collection-time \texttt{ImportError} / \texttt{NameError}, \texttt{NameError} inside \texttt{tools\_implementation.py} or \texttt{data\_model.py}, missing tool methods, or module-level attribute omissions.
\end{itemize}

We define an environment version as \emph{defect-free} iff it has zero level-3 failures, which serves as the filter gate before an env-version is admitted as the base of the next evolution step.

\paragraph{Criticality distribution.} Across the 810 e-commerce failures, we find 292 (36\%) at level 1, 470 (58\%) at level 2, and only 48 (6\%) at level 3, i.e., roughly 0.7\% of all 7{,}315 tests indicate a real tool defect. Airline follows the same shape: 56 / 14 / 2 at levels 1 / 2 / 3 over 72 total failures.

\begin{table}[h]
\centering
\small
\caption{Per-version defect-free rate and failure criticality distribution.
An env-version is defect-free iff it contains zero level-3 (partial-impact)
failures.}
\label{tab:defect-free}
\begin{tabular}{lcccc}
\toprule
\textbf{Version} & \textbf{Defect-free} & \textbf{Cat-1 (no impact)} & \textbf{Cat-2 (corner)} & \textbf{Cat-3 (defect)} \\
\midrule
\multicolumn{5}{l}{\textit{E-commerce} (50 trajectories $\times$ 3 versions = 150 env-versions)} \\
V1 (completion) & 46/50 (92\%)  & 118 & 298 & 46 \\
V2 (saturation) & 50/50 (100\%) &  80 &  77 &  0 \\
V3 (deprecation) & 48/50 (96\%) &  94 &  95 &  2 \\
\textbf{All}    & \textbf{144/150 (96.0\%)} & 292 & 470 & 48 \\
\midrule
\multicolumn{5}{l}{\textit{Airline} (5 trajectories $\times$ 3 versions = 15 env-versions)} \\
V1 (completion)  &  5/5 (100\%) & 41 & 7 & 0 \\
V2 (saturation)  &  5/5 (100\%) & 15 & 4 & 0 \\
V3 (deprecation) &  3/5 (60\%)  &  0 & 3 & 2 \\
\textbf{All}    & \textbf{13/15 (86.7\%)} & 56 & 14 & 2 \\
\bottomrule
\end{tabular}
\end{table}


\begin{table}[h]
  \centering
  \small
  \caption{Breakdown of unit-test failures across all evolved environments.
    We categorize each failing test by its root cause and downstream impact
    on tool usage. Only ${\sim}10\%$ of failures (roughly $1\%$ of all tests)
    affect actual tool behavior.}
  \label{tab:test-failure-breakdown}
  \begin{tabular}{lccl}
    \toprule
    \textbf{Failure Category} & \textbf{\% of Failures} & \textbf{\% of All Tests} & \textbf{Impact on Tool Usage} \\
    \midrule
    Test implementation issues & 51\% & 4.7\% & None --- bug in test, not environment \\
    Corner-case handling       & 23\% & 2.1\% & Minimal --- e.g., invalid-input checks \\
    Real tool issues           & 10\% & 0.9\% & Affects specific tools \\
    Other                      & 16\% & 1.5\% & Varies \\
    \bottomrule
  \end{tabular}
\end{table}

\paragraph{Representative examples.}
\textit{(Level 1)} In \texttt{environment\_48/V2}, the test fixture constructs \texttt{DeliverySlot(slot\_type="morning")}, but the model's \texttt{Literal} only admits \{\texttt{standard}, \texttt{express}, \texttt{weekend}, \texttt{holiday}\}. The tool itself is correct; the fixture uses an out-of-vocabulary label.

\textit{(Level 2)} \texttt{airline\_1/V3} triggers an assertion on a seldom-used response field where the tool returns \texttt{None} instead of a default. Typical user flows are unaffected.

\textit{(Level 3)} \texttt{environment\_15/V1} calls \texttt{timedelta(days=1)} inside \texttt{tools\_implementation.py} without importing \texttt{datetime.timedelta}, producing a collection-time \texttt{NameError} that breaks 36 downstream tests. This env-version is filtered from the pipeline and never serves as a base for V2 generation.

\subsection{Semantic-Intent Alignment Rubric and Scores}
\label{app:semantic-intent}
\label{app:semantic-fidelity}
\label{app:llm-judge}
We evaluate each individual evolution modification using an LLM-as-a-Judge (Claude Sonnet 4.5) that scores three dimensions, each decomposed into three sub-criteria on a 1--5 scale. Score~5 indicates \emph{meets}; 3--4 indicates \emph{partially meets}; $\leq$2 indicates \emph{does not meet}.

\paragraph{Terminology note.} In the current tables, the first dimension is named \textit{Reality}. In the revised paper framing, we interpret this dimension as \textit{Plausibility}: whether the modification is a plausible structured change to a deployed environment.

\begin{table}[h]
\centering
\small
\caption{Scoring rubric for semantic fidelity.}
\label{tab:rubric}
\begin{tabular}{p{0.22\linewidth}p{0.72\linewidth}}
\toprule
\textbf{Sub-criterion} & \textbf{What it checks} \\
\midrule
\multicolumn{2}{l}{\textit{Reality.} Plausibility of the evolution as a real-world change.} \\
\quad Realism        & The evolution reflects a change that could plausibly happen in production (e.g., feature addition, API consolidation, deprecation). \\
\quad Coherence      & The proposal has a clear user story and internally consistent entities/tools. \\
\quad Complexity     & The scope is well-calibrated for a single evolution step---non-trivial yet bounded. \\
\midrule
\multicolumn{2}{l}{\textit{Graph Fidelity.} Alignment between implemented code and the specified graph edit.} \\
\quad Completeness   & All nodes/edges added or removed by the graph edit appear in the implementation. \\
\quad Correctness    & Field types, cardinalities, and tool placements match the graph specification. \\
\quad Consistency    & The new elements follow the existing graph's conventions (naming, FK patterns, edge structure). \\
\midrule
\multicolumn{2}{l}{\textit{Code Fidelity.} Correctness of tool signatures and cross-entity relations.} \\
\quad Model completeness     & All required fields are present on the generated pydantic models. \\
\quad Tool completeness      & All tool methods specified by the graph are implemented. \\
\quad Signature correctness  & Tool parameter and return types match the graph edge definitions. \\
\bottomrule
\end{tabular}
\end{table}

\paragraph{Per-sub-criterion results.} Table~\ref{tab:judge-breakdown} reports mean scores and the fraction of evaluations at each tier. Across all 9 sub-criteria in both domains, \textbf{0\% of evaluations score ``does not meet''}. The tightest sub-criteria are complexity (calibration of scope), graph completeness, and tool completeness, which still sit above 4.4/5 on average.

\begin{table}[h]
\centering
\small
\caption{Per-sub-criterion LLM-as-Judge results.
E-commerce: 10 trajectories $\times$ 3 versions (30 evaluations).
Airline: 5 trajectories $\times$ 3 versions (15 evaluations).}
\label{tab:judge-breakdown}
\begin{tabular}{lcccccc}
\toprule
& \multicolumn{3}{c}{\textbf{E-commerce}} & \multicolumn{3}{c}{\textbf{Airline}} \\
\cmidrule(lr){2-4}\cmidrule(lr){5-7}
\textbf{Sub-criterion} & \textbf{Mean} & \textbf{Meet} & \textbf{Partial} & \textbf{Mean} & \textbf{Meet} & \textbf{Partial} \\
\midrule
Realism             & 4.87 & 90\% & 10\% & 5.00 & 100\% &  0\% \\
Coherence           & 4.80 & 83\% & 17\% & 4.93 &  93\% &  7\% \\
Complexity          & 4.60 & 57\% & 43\% & 4.40 &  40\% & 60\% \\
\midrule
Graph completeness  & 4.40 & 47\% & 53\% & 4.40 &  47\% & 53\% \\
Graph correctness   & 4.77 & 80\% & 20\% & 4.80 &  80\% & 20\% \\
Graph consistency   & 4.73 & 80\% & 20\% & 4.93 &  93\% &  7\% \\
\midrule
Model completeness  & 4.57 & 70\% & 30\% & 4.13 &  33\% & 67\% \\
Tool completeness   & 4.50 & 50\% & 50\% & 4.73 &  73\% & 27\% \\
Signature correctness & 4.57 & 67\% & 33\% & 4.93 &  93\% &  7\% \\
\bottomrule
\end{tabular}
\end{table}

\paragraph{Per-strategy means.} No significant degradation is observed across the three evolution strategies. For e-commerce, Completion/Saturation/Deprecation yield Reality means of 4.67/4.80/4.80, Graph Fidelity 4.78/4.53/4.58, and Code Fidelity 4.33/4.67/4.64 respectively (all $> 4.3$). Airline follows the same pattern.

\subsection{Environment Diversity Metrics}
\label{app:diversity}
\paragraph{Baseline protocol.} We compare \textsc{ProEvolve} against a \textit{random-evolution LLM baseline} that evolves from the same seed environment but without graph-guided transformations. Specifically, the baseline uses the same coding agent with identical seed environment and matched compute budget, but receives no explicit graph edits---it is prompted only with a free-form description of the intended change (e.g., `add a capability'), and generates code directly. Both pipelines run 50 evolution trajectories and produce $V^{(0)} \to V^{(1)}$ pairs.

\paragraph{Metrics.} Let $T_i$ be the set of tools in trajectory~$i$. We report pairwise tool Jaccard distance, unique tool ratio, and tool-frequency entropy. Analogous metrics are computed on entity sets.

\begin{table}[h]
\centering
\small
\caption{Diversity comparison between graph-guided \textsc{ProEvolve}
and a random-evolution LLM baseline, each run over 50 trajectories from
the same e-commerce seed.}
\label{tab:diversity}
\begin{tabular}{lccc}
\toprule
\textbf{Metric} & \textbf{\textsc{ProEvolve}} & \textbf{Random LLM} & \textbf{$\Delta$} \\
\midrule
\multicolumn{4}{l}{\textit{Tool diversity}} \\
Pairwise Jaccard distance $\uparrow$  & \textbf{0.976} & 0.867 & $+0.109$ \\
Unique tool ratio $\uparrow$          & \textbf{0.645} & 0.394 & $+0.251$ \\
Tool-frequency entropy (bits) $\uparrow$ & \textbf{8.07} & 7.66 & $+0.41$ \\
Avg.\ new tools per trajectory        & 10.8 & 21.0 & --- \\
\midrule
\multicolumn{4}{l}{\textit{Entity diversity}} \\
Pairwise Jaccard distance $\uparrow$  & \textbf{0.979} & 0.805 & $+0.174$ \\
Unique entity ratio $\uparrow$        & \textbf{0.606} & 0.336 & $+0.270$ \\
Entity-frequency entropy (bits) $\uparrow$ & \textbf{6.18} & 6.10 & $+0.08$ \\
\bottomrule
\end{tabular}
\end{table}

\begin{table}[h]
  \centering
  \small
  \caption{Environment diversity comparison.
    \textsc{ProEvolve} (graph-guided evolution) vs.\ a \emph{non-graph baseline}
    that receives the same seed specifications but proposes and implements
    evolution without explicit graph transformations.
    All metrics are computed over the 200 e-commerce environment variants.}
  \label{tab:diversity-summary}
  \begin{tabular}{lccc}
    \toprule
    \textbf{Metric}
      & \textbf{\textsc{ProEvolve}}
      & \textbf{No-Graph}
      & \textbf{$\Delta$} \\
    \midrule
    Mean pairwise tool Jaccard dist.\,$\uparrow$
      & 0.976 & 0.867 & \textbf{+0.109} \\
    Mean pairwise entity Jaccard dist.\,$\uparrow$
      & 0.979 & 0.805 & \textbf{+0.174} \\
    Unique tool ratio\,$\uparrow$
      & 0.645 & 0.394 & \textbf{+0.251} \\
    Unique entity ratio\,$\uparrow$
      & 0.606 & 0.336 & \textbf{+0.269} \\
    Tool frequency entropy\,$\uparrow$
      & 8.07  & 7.66  & \textbf{+0.41}  \\
    Entity frequency entropy\,$\uparrow$
      & 6.18  & 6.10  & \textbf{+0.07}  \\
    \bottomrule
  \end{tabular}
\end{table}

\paragraph{Observation.} \textsc{ProEvolve} produces materially more diverse trajectories on every metric despite generating roughly half as many new tools per step (10.8 vs.\ 21.0). Rather than repeatedly reusing a small set of common tool patterns, the graph-guided pipeline induces structurally diverse environment variants.

\subsection{Task Quality Analysis}
\label{app:task-quality}

We evaluate the intrinsic quality of generated task sandboxes using an LLM-as-a-Judge (Claude Sonnet 4.5) that scores each task along four dimensions on a 1--5 scale. We sample 200 task sandboxes balanced across difficulty levels (easy, medium, hard) and evolution primitives (seed, completion, saturation, deprecation). Table~\ref{tab:task-rubric} defines the scoring rubrics; Table~\ref{tab:task-quality-scores} reports aggregate results.

\begin{table}[h]
\centering
\small
\caption{Scoring rubric for task quality evaluation. Score 5 = fully meets; 3--4 = partially meets; $\leq$2 = does not meet.}
\label{tab:task-rubric}
\begin{tabular}{p{0.22\linewidth}p{0.72\linewidth}}
\toprule
\textbf{Dimension} & \textbf{What it checks} \\
\midrule
\multicolumn{2}{l}{\textit{Scenario Coherence.} Does the sequence of states form a realistic, logical customer journey?} \\
\quad Narrative arc    & States follow a natural progression; no random or disconnected actions. \\
\quad Transition logic & Each state follows plausibly from its predecessor without forced jumps. \\
\midrule
\multicolumn{2}{l}{\textit{Difficulty Calibration.} Does the task complexity match the claimed difficulty level?} \\
\quad State count      & Easy: 5--6 states; medium: 6--8; hard: 8--10. \\
\quad Operation mix    & Easy: mostly reads; medium: reads + simple writes; hard: multi-entity read/write chains. \\
\midrule
\multicolumn{2}{l}{\textit{State Transition Feasibility.} Is each transition (State $N \to N{+}1$) executable given available information?} \\
\quad Data handoff     & Each state's outputs provide sufficient inputs for the next state. \\
\quad No impossible jumps & The customer is never required to possess information not yet revealed. \\
\midrule
\multicolumn{2}{l}{\textit{Init Environment Sufficiency.} Are prerequisite entities properly materialized?} \\
\quad Entity coverage  & All non-pickable entities (orders, carts, reservations) referenced by the task are initialized. \\
\quad Cross-references & Shared identifiers and FK links across entities are consistent. \\
\bottomrule
\end{tabular}
\end{table}

\paragraph{Judging conventions.}
Authentication after browsing is acceptable (browsing does not require \texttt{user\_id}). User accounts and product catalogs pre-exist in the base database and need not be explicitly initialized. Empty \texttt{init\_actions} is acceptable for tasks that create entities at runtime (e.g., search $\to$ cart $\to$ checkout).

\begin{table}[h]
\centering
\small
\caption{Task quality scores (1--5 scale) judged by LLM-as-a-Judge across sampled sandboxes.}
\label{tab:task-quality-scores}
\begin{tabular}{lcc}
\toprule
Dimension & E-com. & Airline \\
\midrule
Scenario Coherence           & 4.75 & 4.81 \\
Difficulty Calibration       & 5.00 & 5.00 \\
State Transition Feasibility & 4.50 & 4.06 \\
Prerequisite Sufficiency     & 4.62 & 4.88 \\
\bottomrule
\end{tabular}
\end{table}

Across all four dimensions, zero evaluations score ``Fails'' ($\leq 2$), confirming that graph-grounded task generation produces coherent, well-calibrated, and executable sandboxes. The tightest dimension is \emph{Difficulty Calibration} (mean 4.48), reflecting occasional boundary cases where a task's state count slightly deviates from the expected range for its difficulty label---these are acceptable under the rubric (score 4) and do not affect executability.

\section{Full Agent Evaluation Results}
\label{app:agent-results}
This section provides detailed agent-evaluation results supporting the compact diagnostics reported in Sec.~\ref{subsec:env_evolution}. The main paper reports aggregated trends by evolution primitive, performance--efficiency tradeoffs, and failure-mode breakdowns; here we provide the full per-model, per-version, and per-adaptation results for transparency and reproducibility.

\paragraph{Evaluation setup.} We use \texttt{Litellm} \cite{agarwal2024litllm} to make LLM API calls, and estimate the corresponding API cost. To evaluate the generated benchmark, we testify with \texttt{gpt-5-2025-08-07}~\cite{singh2025openai}, \texttt{Gemini 2.5-Pro}~\cite{comanici2025gemini}, \texttt{claude-opus-4-5-20251101}~\cite{claude}, \texttt{Qwen3-235B-A22B-Thinking-2507}~\cite{qwen3}, and \texttt{DeepSeek-V3.2}~\cite{liu2025deepseek}. 
Like $\tau^2$ ~\cite{barres2025tau2}, we instruct an LLM-powered customer simulator \texttt{claude-opus-4-5-20251101} to act as the customer to chat with the agent given the state-wise instructions.

\subsection{Main-Domain Detailed Results: E-Commerce}
\label{app:ecom-results}
\begin{table*}[t]
\centering

\label{tab:finalset_results}
\resizebox{\textwidth}{!}{%
\begin{tabular}{l|ccc|ccc|ccc|ccc|ccc}
\hline

Evolving Trajectory & & $\mathcal{G}^{(0)}$ & \multicolumn{2}{c}{$\xrightarrow{\Delta^{\text{comp}}}$} & $\mathcal{G}^{(1)}$ & \multicolumn{2}{c}{$\xrightarrow{\Delta^{\text{sat}}}$}  &  $\mathcal{G}^{(2)}$ & \multicolumn{2}{c}{$\xrightarrow{\Delta^{\text{dep}}}$}& $\mathcal{G}^{(3)}$ &  & \multicolumn{3}{c}{\textbf{Overall}}\\

\hline 
{Strategy} & $\mathcal{C}^{(0)}$ & $\overline{T}$ & $\overline{N}_{\mathrm{tool}}$ & $\mathcal{C}^{(1)}$ & $\overline{T}$ & $\overline{N}_{\mathrm{tool}}$ & $\mathcal{C}^{(2)}$ & $\overline{T}$ & $\overline{N}_{\mathrm{tool}}$ & $\mathcal{C}^{(3)}$ & $\overline{T}$ & $\overline{N}_{\mathrm{tool}}$ & $\mu_{C}$ & $\overline{T}$ & $\overline{N}_{\mathrm{tool}}$ \\

\hline
\hline
\multicolumn{16}{c}{\textit{GPT-5}} \\
\hline
Baseline & 0.564 & 8.1 & 8.1 & 0.646 & 10.3 & 12.8 & 0.771 & 13.2 & 14.9 & 0.454 & 8.7 & 20.6 & 0.609 & 10.1 & 14.1 \\
Trace Context  & 0.562 & 8.9 & 10.2 & 0.667 & 9.4 & 13.5 & 0.786 & 11.6 & 14.4 & 0.407 & 6.5 & 7.6 & 0.606 & 9.1 & 11.4 \\
Summary Context & 0.598 & 12.3 & 12.2 & 0.537 & 10.1 & 11.1 & 0.677 & 11.8 & 13.1 & 0.530 & 9.2 & 10.3 & 0.585 & 10.9 & 11.7 \\
\hline
\hline
\multicolumn{16}{c}{\textit{Claude-Opus-4.5}} \\
\hline
Baseline & 0.486 & 6.3 & 4.7 & 0.463 & 5.5 & 4.8 & 0.623 & 7.3 & 5.9 & 0.369 & 4.5 & 4.4 & 0.485 & 5.9 & 4.9 \\
Trace Context  & 0.542 & 6.7 & 5.3 & 0.540 & 6.9 & 5.7 & 0.540 & 6.8 & 5.5 & 0.358 & 4.1 & 2.3 & 0.495 & 6.1 & 4.7 \\
Summary Context& 0.438 & 6.2 & 9.6 & 0.558 & 6.7 & 17.3 & 0.501 & 6.3 & 6.2 & 0.336 & 4.7 & 6.6 & 0.458 & 6.0 & 9.9 \\
\hline
\hline
\multicolumn{16}{c}{\textit{DeepSeek-V3.2}} \\
\hline
Baseline & 0.529 & 7.3 & 6.7 & 0.517 & 7.4 & 6.9 & 0.490 & 8.5 & 6.7 & 0.462 & 9.0 & 7.3 & 0.500 & 8.1 & 6.9 \\
Trace Context  & 0.681 & 11.5 & 12.3 & 0.538 & 7.8 & 8.4 & 0.565 & 8.3 & 8.3 & 0.373 & 6.1 & 6.9 & 0.539 & 8.4 & 9.0 \\
Summary Context& 0.577 & 8.1 & 9.7 & 0.693 & 11.4 & 15.8 & 0.479 & 8.1 & 9.7 & 0.516 & 9.8 & 13.2 & 0.566 & 9.3 & 12.1 \\
\hline
\hline
\multicolumn{16}{c}{\textit{Qwen3-235B}} \\
\hline
Baseline & 0.588 & 10.4 & 9.5 & 0.633 & 12.9 & 13.1 & 0.460 & 9.0 & 8.1 & 0.364 & 7.6 & 7.4 & 0.511 & 10.0 & 9.5 \\
Trace Context & 0.507 & 9.0 & 8.1 & 0.534 & 8.6 & 7.7 & 0.484 & 7.5 & 6.8 & 0.380 & 7.3 & 6.9 & 0.476 & 8.1 & 7.4 \\
Summary Context & 0.411 & 5.7 & 8.5 & 0.661 & 11.0 & 14.1 & 0.535 & 6.7 & 7.1 & 0.193 & 3.1 & 2.6 & 0.450 & 6.6 & 8.1 \\
\hline
\hline
\multicolumn{16}{c}{\textit{Gemini-2.5-Pro}} \\
\hline
Baseline & 0.427 & 7.3 & 3.6 & 0.572 & 8.9 & 5.1 & 0.543 & 8.0 & 3.9 & 0.453 & 8.1 & 3.9 & 0.499 & 8.1 & 4.2 \\
Trace Context & 0.412 & 6.5 & 3.2 & 0.482 & 7.1 & 4.7 & 0.480 & 8.3 & 4.3 & 0.333 & 5.7 & 3.0 & 0.426 & 6.9 & 3.8 \\
Summary Context& 0.441 & 7.4 & 4.0 & 0.451 & 5.7 & 3.8 & 0.456 & 6.7 & 3.3 & 0.437 & 7.7 & 5.0 & 0.446 & 6.8 & 4.0 \\
\hline

\end{tabular}%

}
\caption{\textbf{Evolving-trajectory evaluation across environment transformations.}
We report per-version task success rate $\mathcal{C}^{(k)}$, average turns $\overline{T}$, and average tool calls $\overline{N}_{\text{tool}}$ for a fixed evolving trajectory $\mathcal{G}^{(0)}\!\rightarrow\!\mathcal{G}^{(1)}\!\rightarrow\!\mathcal{G}^{(2)}\!\rightarrow\!\mathcal{G}^{(3)}$ generated by applying three evolution strategies (arrows): $\Delta^{\mathrm{comp}}$, $\Delta^{\mathrm{sat}}$, and $\Delta^{\mathrm{dep}}$.
Each block compares adaptation strategies (Baseline, Trace Context, Summary Context) for a given model, and the \textbf{Overall} columns summarize mean completeness $\mu_{\mathcal{C}}$ and average interaction cost aggregated over versions.}
\label{table:episode_results}
\end{table*}

E-commerce is the main large-scale diagnostic domain used in Sec.~\ref{subsec:env_evolution}. Table~\ref{table:episode_results} reports the full breakdown behind the stage-wise aggregate results in the main paper: five models, three adaptation strategies, and four environment versions along the trajectory
\[
G^{(0)} \xrightarrow{\Delta_{\mathrm{comp}}} G^{(1)}
\xrightarrow{\Delta_{\mathrm{sat}}} G^{(2)}
\xrightarrow{\Delta_{\mathrm{dep}}} G^{(3)}.
\]
We report state-wise success $\mathcal{C}^{(k)}$, average turns $\bar{T}$, and average tool calls $\bar{N}_{\mathrm{tool}}$ for each version. These detailed results are intended as supplementary evidence rather than a main-paper leaderboard.

The results show substantial variation across both environment versions and agents. For example, GPT-5 with trace context increases from $0.562$ at $\mathcal{G}^{(0)}$ to $0.786$ at $\mathcal{G}^{(2)}$, but drops to $0.407$ after deprecation at $\mathcal{G}^{(3)}$. Other agents exhibit different trajectories and tool-use patterns, indicating that structured environment changes induce heterogeneous performance and efficiency responses. These full results provide supplementary evidence for the main-paper claim that different evolution primitives create distinct diagnostic pressures.

\subsection{Cross-Domain Pilot Results: Airline}
\label{app:airline-results}

We use airline booking as a smaller cross-domain pilot to examine whether the generated environment trajectories are executable and evaluable beyond e-commerce. Unlike e-commerce, which serves as the main large-scale diagnostic benchmark, airline is not intended as a full cross-domain leaderboard. We therefore report compact per-version results and use them as supplemental evidence for the transfer of the benchmark-construction pipeline to another structured tool-calling domain.

\begin{table*}[h]
\centering
\resizebox{\textwidth}{!}{%
\begin{tabular}{l|ccc|ccc|ccc|ccc|cc}
\hline

Evolving Trajectory & & $\mathcal{G}^{(0)}$ & \multicolumn{2}{c}{$\xrightarrow{\Delta^{\text{comp}}}$} & $\mathcal{G}^{(1)}$ & \multicolumn{2}{c}{$\xrightarrow{\Delta^{\text{sat}}}$}  &  $\mathcal{G}^{(2)}$ & \multicolumn{2}{c}{$\xrightarrow{\Delta^{\text{dep}}}$}& $\mathcal{G}^{(3)}$ &  & \multicolumn{2}{c}{\textbf{Overall}}\\

\hline 
{Strategy} & $\mathcal{C}^{(0)}$ & $\overline{T}$ & $\overline{N}_{\mathrm{tool}}$ & $\mathcal{C}^{(1)}$ & $\overline{T}$ & $\overline{N}_{\mathrm{tool}}$ & $\mathcal{C}^{(2)}$ & $\overline{T}$ & $\overline{N}_{\mathrm{tool}}$ & $\mathcal{C}^{(3)}$ & $\overline{T}$ & $\overline{N}_{\mathrm{tool}}$ & $\mu_{\mathcal{C}}$ & Cost \\

\hline
\hline
\multicolumn{15}{c}{\textit{Claude-Opus-4.5}} \\
\hline
Baseline          & 0.58 & 11 & 12 & 0.49 & 7 & 9  & 0.54 & 9 & 10 & 0.27 & 9  & 12 & 0.47 & \$0.68 \\
Trace Context              & 0.50 & 9  & 18 & 0.35 & 7 & 10 & 0.58 & 8 & 12 & 0.43 & 10 & 11 & 0.47 & \$0.70 \\
Summary Context       & 0.60 & 11 & 22 & 0.29& 7 & 14 & 0.25 & 6 & 14 & 0.20 & 9  & 16 & 0.34 & \$1.02 \\
\hline
\hline
\multicolumn{15}{c}{\textit{DeepSeek-V3.2}} \\
\hline
Baseline          & 0.35 & 13 & 16 & 0.34 & 12 & 10 & 0.53& 14 & 22 & 0.49& 14 & 13 & 0.43 & \$0.21 \\
Trace Context              & 0.60 & 16 & 18 & 0.33 & 8  & 9  & 0.56 & 9  & 11 & 0.33 & 9  & 8  & 0.46 & \$0.14 \\
Summary Context        & 0.60  & 11 & 12 & 0.40 & 10 & 16 & 0.39  & 11 & 15 & 0.47 & 12 & 13 & 0.46  & \$0.17 \\
\hline
\hline
\multicolumn{15}{c}{\textit{Qwen3-235B}} \\
\hline
Baseline          & 0.33 & 9  & 11 & 0.20  & 9  & 7  & 0.33 & 10 & 9  & 0.31 & 11 & 13 & 0.29 & \$0.07 \\
Trace Context         & 0.43  & 9  & 8  & 0.30 & 12 & 10 & 0.37  & 9  & 11 & 0.35  & 12 & 15 & 0.36 & \$0.08 \\
Summary Context        & 0.23 & 9  & 11 & 0.41 & 12 & 11 & 0.23 & 10 & 12 & 0.47  & 11 & 14 & 0.34 & \$0.09 \\
\hline

\end{tabular}%
}
\caption{\textbf{Airline domain: evolving-trajectory evaluation.}
We report per-version consecutive success rate $\mathcal{C}^{(k)}$, average turns $\overline{T}$, and average tool calls $\overline{N}_{\text{tool}}$ for a fixed evolving trajectory $\mathcal{G}^{(0)}\!\rightarrow\!\mathcal{G}^{(1)}\!\rightarrow\!\mathcal{G}^{(2)}\!\rightarrow\!\mathcal{G}^{(3)}$ generated by applying three evolution strategies: $\Delta^{\mathrm{comp}}$, $\Delta^{\mathrm{sat}}$, and $\Delta^{\mathrm{dep}}$.
The \textbf{Overall} columns report mean success $\mu_{\mathcal{C}}$ and estimated API cost per task. }
\label{tab:airline_trajectory}
\end{table*}

Table~\ref{tab:airline_trajectory} shows that the airline environments are executable and induce nontrivial evaluation signals across versions. The trends are model- and probe-dependent: for Opus-4.5, Summary context substantially reduces overall success and increases tool usage, whereas trace context is more stable; for DeepSeek-V3.2, Memory and Reflection yield similar aggregate success but different interaction budgets; Qwen3-VL remains lower overall but shows variation across versions. These results support airline as a cross-domain pilot for the benchmark-construction pipeline, while the main large-scale diagnostic conclusions remain based on e-commerce.

\subsection{Context Probe Details}
\label{app:replay-probes}

\paragraph{Purpose.}
We use context conditions as diagnostic probes for prior-version experience, not as proposed adaptation methods. They test whether information from earlier environment versions remains useful, becomes irrelevant, or becomes misleading as tools, schemas, and access paths change.

\paragraph{Context conditions.}
We compare three conditions. \textbf{No Prior Context} solves each task independently without prior-version experience. \textbf{Trace Context} provides the agent with previous interaction traces from earlier environment versions. \textbf{Summary Context} provides distilled natural-language summaries of prior experience instead of raw traces.

\paragraph{Implementation.}
For each environment version, context information is constructed only from previous versions in the same evolution trajectory. Thus, context never includes future-version information. The context is prepended to the agent prompt before the current task begins. We keep the task instruction, simulator protocol, and evaluation metric unchanged across context conditions.

\paragraph{Interpretation.}
Context probes are useful as a diagnostic because prior-version experience is not guaranteed to remain valid under structured evolution. Trace context may help when changes preserve useful access paths, but it may also become stale after deprecation or access-path reorganization. Summary context can reduce prompt length, but distilled summaries may overgeneralize previous tool availability or schema assumptions.

\subsection{Failure-Mode Analysis}
\label{app:failure-modes}

We decompose agent failures across environment versions and context conditions to understand \emph{how} agents fail under structured evolution, not just how often. For each failed state in a sampled trajectory, we classify the root cause into one of five categories defined below.

\paragraph{Failure taxonomy.}
We define five failure patterns from a tool-calling perspective, ordered by the stage of the agent pipeline at which the failure occurs:

\begin{table}[h]
\centering
\small
\caption{Failure-mode taxonomy for tool-calling agents.}
\label{tab:failure-taxonomy}
\begin{tabular}{p{0.24\linewidth}p{0.70\linewidth}}
\toprule
\textbf{Failure Mode} & \textbf{Definition} \\
\midrule
Missed tool discovery & The agent does not know a relevant tool exists. It tells the user ``I can't do that'' or ``I don't have access'' when an appropriate tool is available. Failure at the discovery/awareness level. \\
Invalid parameters & The agent identifies and calls the correct tool but passes wrong arguments---wrong entity IDs, malformed field values, or incorrect types. The tool errors out or returns incorrect results. \\
Missing invocation & The agent understands what needs to happen and knows the tool exists, but does not proactively call it. It waits passively, provides text explanations, or lets the user drive the next step. Failure in initiative/sequencing, not knowledge. \\
Unrecovered error & A tool call returns an explicit error (e.g., ``product not found''), and the agent fails to adapt---it retries with the same bad parameters, gives up entirely, or tells the user to try later instead of attempting an alternative approach. \\
Infeasible expectation & The expected state requires an action that genuinely cannot be accomplished with the available tool set (e.g., triggering warehouse processing or initiating a live chat). This reflects a benchmark design boundary rather than an agent failure. \\
\bottomrule
\end{tabular}
\end{table}

\begin{figure}[h]
\centering
\includegraphics[width=\textwidth]{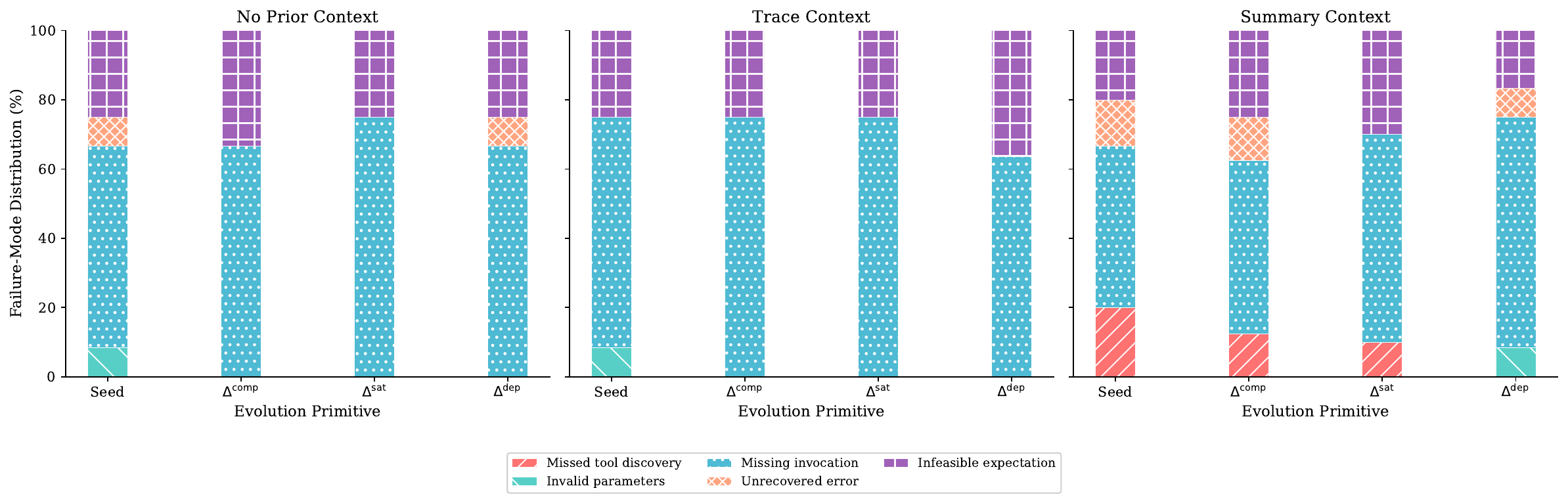}
\caption{Failure-mode distribution across environment versions and context conditions (agent: Claude-Opus-4.5). Each bar decomposes sampled failed states by root cause. \emph{Missing invocation} dominates under No Prior Context and Trace Context, while \emph{Summary Context} shifts failures toward \emph{missed tool discovery}---indicating that distilled summaries can actively mislead agents about current tool availability. \emph{Unrecovered error} increases under $\Delta^{\text{dep}}$, reflecting retry loops on deprecated tools.}
\label{fig:failure-modes-all}
\end{figure}

\paragraph{Results across context conditions.}
Figure~\ref{fig:failure-modes-all} reports the failure-mode distribution across all four environment versions (Seed, $\Delta^{\text{comp}}$, $\Delta^{\text{sat}}$, $\Delta^{\text{dep}}$) under three context conditions (No Prior Context, Trace Context, Summary Context). Several patterns emerge:

\begin{itemize}[leftmargin=*, nosep]
    \item \textbf{Missing invocation dominates across all settings.} Under No Prior Context and Trace Context, missing invocation accounts for $\sim$55--65\% of failures regardless of evolution primitive, indicating that the primary bottleneck is agent initiative---knowing a tool exists but failing to act on it proactively.

    \item \textbf{Summary Context shifts failures toward missed tool discovery.} Under Summary Context, missed tool discovery (red) becomes substantially more prevalent, particularly at the Seed and $\Delta^{\text{comp}}$ versions. This suggests that distilled summaries from prior versions can actively mislead agents about tool availability---the agent internalizes outdated beliefs about what tools exist rather than consulting the current tool list.

    \item \textbf{Infeasible expectations are stable across versions.} The proportion of infeasible expectation failures (purple) remains roughly constant ($\sim$20--25\%) across evolution primitives and context conditions, confirming that this category reflects inherent benchmark boundaries rather than agent-side degradation.

    \item \textbf{Unrecovered errors spike under deprecation.} Under $\Delta^{\text{dep}}$, unrecovered error (orange) increases compared to other versions, particularly under Summary Context. Agents attempt deprecated tools, receive errors, and fail to recover---consistent with the retry-loop behavior observed in Table~2 (e.g., GPT-5's tool calls spiking to 20.6 under $\Delta^{\text{dep}}$).

    \item \textbf{Invalid parameters remain a minor mode.} Across all conditions, invalid parameters (cyan) account for $<$10\% of failures, suggesting that once agents identify and invoke a tool, they generally construct valid arguments.
\end{itemize}

\paragraph{Implications.}
The failure decomposition reveals that structured evolution primarily stresses \emph{tool discovery} and \emph{initiative}, not argument construction. Context strategies interact with these failure modes in distinct ways: Trace Context preserves the baseline failure distribution (suggesting raw traces neither help nor hurt at the categorical level), while Summary Context introduces a new failure pathway by encouraging agents to rely on outdated tool beliefs rather than re-discovering capabilities in the current version.

\section{Broader Impact}
\label{app:impact}
\textbf{Positive Contributions.}
This work advances evaluation methodology for LLM-powered tool-calling agents by introducing a framework for constructing benchmarks with structured, version-linked environment evolution. We expect several positive impacts.

\begin{enumerate}
    \item \textbf{Improved diagnostic evaluation.} 
    By generating executable environments where tools, schemas, and data access change coherently across versions, \textsc{ProEvolve} helps expose failure modes that may be missed by static benchmarks, such as reliance on deprecated tools, failure to discover newly introduced capabilities, or inefficient recovery after access-path changes.
    
    \item \textbf{More systematic benchmark construction.} 
    The graph-based formulation provides a controlled way to create environment trajectories and version-linked task sandboxes. This supports more transparent evaluation of tool-calling agents under structured changes, rather than relying only on fixed toolsets or independently sampled environments.
    
    \item \textbf{Lower cost of benchmark generation.} 
    Automated environment and task generation can help researchers create diverse diagnostic scenarios with less manual effort, making it easier to study agent behavior under changing interfaces and capabilities.
\end{enumerate}

\textbf{Potential Risks and Mitigation.}
\begin{enumerate}
    \item \textbf{Over-reliance on generated benchmarks.} 
    Although \textsc{ProEvolve} improves coverage beyond static benchmarks, generated environments may still contain artifacts from the graph abstraction, LLM generation, or user simulator. Performance on this benchmark should therefore be interpreted as diagnostic evidence, not as a guarantee of robustness in deployed systems. We recommend using it alongside human-designed tests, real logs where available, and domain-specific evaluations.
    
    \item \textbf{Incomplete domain coverage.} 
    Our framework targets structured tool-calling environments with explicit schemas, data entities, and tool-enabled transitions. Applying it to higher-stakes or less structured domains, such as healthcare, software engineering, or open-ended web/OS control, would require additional domain expertise, safety checks, and human validation.
    
    \item \textbf{Misuse for adversarial evaluation.} 
    The framework could be adapted to generate challenging or adversarial environment variants. While our goal is diagnostic robustness evaluation, such use should be accompanied by clear reporting of generation procedures, validation criteria, and failure categories to avoid misleading claims about agent capability.
\end{enumerate}

Overall, we view \textsc{ProEvolve} as a tool for improving the transparency and diagnostic value of tool-calling agent evaluation. Its main benefit is to help researchers identify brittle behavior under structured environment changes before such agents are deployed in settings where tools and schemas evolve over time.

\section{Version-Linked Task Sandbox Generation}
\label{app:task-generation}
\label{app:task_generation_example}
To explicitly illustrate the task generation pipeline described in Section 4.3, we present a simplified, representative walkthrough of generating a `Product Exchange' task within the e-commerce environment.

Because the sampled task subgraph is drawn from the current environment version $\mathcal{G}^{(k)}$, the resulting task only references tools, entities, and reachable relations available in that version.

\textbf{Step 1: Subgraph Sampling}
The framework first samples a connected subgraph $\mathcal{G}_{\tau} \subseteq \mathcal{G}$ that defines the structural scope of the task. This subgraph ensures that a valid path exists between the user's starting state (Authentication) and the desired outcome (Exchange).

\begin{itemize}
    \item Sampled Nodes ($\mathcal{V}_{\tau}$): \texttt{User} $\rightarrow$ \texttt{Order} $\rightarrow$ \texttt{Product} $\rightarrow$ \texttt{ExchangeRequest}.
    \item Sampled Edges/Tools ($\mathcal{E}_{\tau}$):
    \begin{itemize}
        \item \texttt{authenticate\_user()} (connects Session to User)
        \item \texttt{get\_order\_by\_id()} (connects User to Order History)
        \item \texttt{initiate\_exchange()} (connects Order to ExchangeRequest)
    \end{itemize}
    \item Goal Synthesis ($g_{\tau}$): The LLM formulates a goal realizable within $\mathcal{G}_{\tau}$: \textit{"Exchange the delivered grey Burkini for the correct purple variant."}
    \item Scenario Description ($s_{\tau}$): \textit{You are Susan Morales. You ordered a purple Meijunter Burkini (size M) but received grey. You need to verify the order details and initiate an exchange.}
\end{itemize}

\textbf{Step 2: Sandbox Materialization}
The framework instantiates the specific entities required by $\mathcal{G}_{\tau}$ to ensure the task is executable:

\begin{itemize}
\item A \texttt{User} entity is created: \texttt{Susan Morales} (ID: \texttt{user\_089}).
\item An \texttt{Order} entity is injected (\texttt{ORD-2024-089-001}) with a specific anomaly: the \texttt{items} list contains the \texttt{grey} variant, while the user's intent is \texttt{purple}.
\item The \texttt{Product} database is populated with \texttt{B0861GWJV2} (Meijunter Burkini) to ensure variant data exists.
\end{itemize}

\paragraph{Prerequisite composition.}
When a task requires multiple prerequisites, we jointly instantiate the prerequisite nodes and edges induced by the sampled subgraph. Shared identifiers and cross-entity references are enforced so that all required tool transitions are executable under the corresponding environment version. For example, an exchange task requires a valid user, a delivered order, a product variant mismatch, and an exchange-capable order state; these entities are materialized together rather than independently.

\textbf{Step 3: Agentic Walk Execution}
To generate the multi-turn dialogue, a simulator executes a reference trajectory by expanding along $\mathcal{G}_{\tau}$. At each turn $t$, the simulator identifies the \textit{frontier set} (newly reachable nodes) and generates a state-wise customer instruction $u_t^*$.

\begin{itemize}
\item Turn 1 (Authentication):
\begin{itemize}
\item Frontier: The simulator identifies the \texttt{User} node is reachable via authentication.
\item Generated Utterance ($u_1^*$): \textit{"Hi, I need to check on an order. My email is susan.morales.user\_089@example.com."}
\item Action ($a_1$): Agent calls \texttt{authenticate\_user(email="...")}. 
\end{itemize}
\item Turn 2 (Order Verification):
\begin{itemize}
\item Frontier: With the user authenticated, specific \texttt{Order} nodes become reachable.
\item Generated Utterance ($u_2^*$): \textit{"I ordered a purple Burkini but received grey. The order number is ORD-2024-089-001. Can you check the details?"}
\item Action ($a_2$): Agent calls \texttt{get\_order\_by\_id(order\_id="ORD-2024-089-001")}. 
\end{itemize}
\item Turn 3 (Exchange Initiation):
\begin{itemize}
\item Frontier: With the order details exposed (confirming the grey variant was shipped), the \texttt{ExchangeRequest} node becomes reachable via \texttt{initiate\_exchange()}.
\item Generated Utterance ($u_3^*$): \textit{"Yes, that confirms it was the wrong color. Please start an exchange for the purple size M."}
\item Action ($a_3$): Agent calls \texttt{initiate\_exchange(order\_id=..., new\_variant="purple\_tag m")}. 
\end{itemize}
\end{itemize}
\section{Licenses for Existing Assets}
\label{app:license}
We list the licenses for all external assets used in this work:

\begin{itemize}
    \item \textbf{WebShop} \cite{yao2022webshop}: Released under the MIT License. We use the product catalog (1,000 products) as seed data for the e-commerce environment.
    \item \textbf{$\tau^2$-bench} \cite{barres2025tau2}: Released under the MIT License. We adapt the airline booking domain schema and seed environment specification as the starting point for our airline evolution trajectories.
    \item \textbf{LiteLLM} \cite{agarwal2024litllm}: Released under the MIT License. Used as the unified interface for making LLM API calls during evaluation.
\end{itemize}

All commercial LLM APIs (GPT-5, Claude-Opus-4.5, Gemini-2.5-Pro, DeepSeek-V3.2, Qwen3-235B) are accessed through their respective providers' standard API agreements. Our generated benchmark assets (environment versions, task sandboxes, and evaluation code) will be released under the MIT License upon acceptance.

\clearpage

\section{Illustrative Examples of Structured Environment Evolution}
\label{app:evolution-examples}
\label{app:env-evolution-examples}
We provide representative evolution contexts in e-commerce and airline booking to illustrate how semantic evolution intents are translated into graph edits and executable environment changes. These examples cover the three primitives used in the main paper: Completion, Saturation, and Deprecation.

\begin{table}[h]
\centering
\small
\caption{Representative evolution examples used in Appendix~\ref{app:evolution-examples}.}
\begin{tabular}{lll}
\toprule
Domain & Primitive & Example capability/change \\
\midrule
E-commerce & Completion & Competitor price monitoring / alerts \\
Airline & Completion & Redeem miles for flight upgrade \\
E-commerce & Saturation & Order refund summary shortcut \\
Airline & Saturation & Reservation flight status shortcut \\
E-commerce & Deprecation & Cart service unavailable \\
Airline & Deprecation & Direct-flight search deprecated \\
\bottomrule
\end{tabular}
\end{table}

\subsection{Completion strategy under e-commerce domain}
\fbox{
\begin{minipage}{0.97\linewidth}
\textbf{V1 Evolution Context} \hfill \textbf{Strategy:} \texttt{completion}

\vspace{4pt}
\textbf{Task: Competitor Price Monitoring with Auto-Adjust Cart Alerts}

\textbf{User Story.}
\emph{``As a shopper, I want to set target prices for products in my cart and across competitor websites, and be notified when the best price becomes available---either via a price drop on this website or a better deal at other Website.''}

\vspace{4pt}
\textbf{Why Not Supported.}
\begin{enumerate}\itemsep2pt
    \item \textbf{No \texttt{PriceHistory} entity:} only current pricing is stored (\texttt{Product.price}), with no historical tracking.
    \item \textbf{No \texttt{PriceAlert} entity:} users cannot define target prices, alert conditions, or notification preferences.
    \item \textbf{No \texttt{CompetitorPrice} entity:} competitor pricing is neither modeled nor comparable.
    \item \textbf{No monitoring or scheduling tools:} existing tools are transactional, with no background tracking.
    \item \textbf{No general notification mechanism:} notifications are task-specific (e.g., returns), not reusable.
    \item \textbf{No user price preference storage:} price sensitivity and alert defaults are not persisted.
    \item \textbf{No price analytics APIs:} no comparison against historical, competitor, or threshold-based signals.
\end{enumerate}

\vspace{4pt}
\textbf{Required Capabilities Missing.}

\textbf{Tools:}
\begin{itemize}\itemsep2pt
    \item \texttt{create\_price\_alert(user\_id, product\_id, target\_price, alert\_type, include\_competitors?, competitor\_list?)}
    \item \texttt{get\_price\_alerts(user\_id, status?, product\_id?)}
    \item \texttt{delete\_price\_alert(user\_id, alert\_id)}
    \item \texttt{get\_price\_history(product\_id, days?, granularity?)}
    \item \texttt{get\_competitor\_prices(product\_id)}
    \item \texttt{get\_price\_insights(product\_id)}
    \item \texttt{trigger\_price\_notification(alert\_id, trigger\_type, current\_price, trigger\_source?)}
\end{itemize}

\textbf{Entities:}
\begin{itemize}\itemsep2pt
    \item \texttt{PriceAlert}: alert\_id, user\_id, product\_id, target\_price, alert\_type, status, include\_competitors, competitor\_list, created\_at, triggered\_at, notification\_method
    \item \texttt{PriceHistory}: record\_id, product\_id, price, recorded\_at, source, currency
    \item \texttt{CompetitorPrice}: competitor\_price\_id, product\_id, competitor\_name, competitor\_url, competitor\_price, last\_updated, in\_stock, shipping\_cost
    \item \texttt{UserPricePreferences}: user\_id, default\_alert\_threshold\_percentage, preferred\_notification\_method, notification\_frequency, tracked\_competitors
\end{itemize}
\end{minipage}
}

\subsection{Completion strategy under airline domain}
\fbox{
\begin{minipage}{0.97\linewidth}
\textbf{V1 Evolution Context} \hfill \textbf{Strategy:} \texttt{completion}

\vspace{4pt}
\textbf{Task: Redeem Miles for Flight Upgrade}

\textbf{User Story.}
\emph{``As a frequent flyer with accumulated miles, I want to redeem my miles to upgrade an existing economy reservation to business class, so I can enjoy a more comfortable flight without paying the full cash difference.''}

\vspace{4pt}
\textbf{Why Not Supported.}
\begin{enumerate}\itemsep2pt
    \item \textbf{No mileage balance tracking:} the \texttt{User} entity has no field for accumulated miles (e.g., \texttt{miles\_balance}); \texttt{User.membership} only tracks tier.
    \item \textbf{No mileage transaction history:} there is no entity recording how miles are earned, redeemed, transferred, or expired.
    \item \textbf{No mileage earning rules:} the system lacks configuration for earning miles based on distance, cabin class, fare type, or tier multipliers.
    \item \textbf{No redemption rate configuration:} there is no data defining miles required for cabin upgrades across routes or distances.
    \item \textbf{No upgrade availability check with miles:} existing APIs (e.g., \texttt{update\_reservation\_flights}) only support cash-based changes.
    \item \textbf{No miles redemption execution:} no tool exists to deduct miles and apply an upgrade to an existing reservation.
    \item \textbf{No miles expiration tracking:} the system cannot track earning dates or expiration policies for loyalty miles.
\end{enumerate}

\vspace{4pt}
\textbf{Required Capabilities Missing.}

\textbf{Tools:}
\begin{itemize}\itemsep2pt
    \item \texttt{get\_user\_miles\_balance(user\_id)}
    \item \texttt{get\_upgrade\_miles\_cost(reservation\_id, target\_cabin)}
    \item \texttt{check\_upgrade\_availability(reservation\_id, target\_cabin)}
    \item \texttt{redeem\_miles\_for\_upgrade(user\_id, reservation\_id, target\_cabin)}
\end{itemize}

\textbf{Entities:}
\begin{itemize}\itemsep2pt
    \item \texttt{MilesAccount}: user\_id, current\_balance, lifetime\_miles, pending\_miles, miles\_expiring\_soon, expiration\_date
    \item \texttt{MilesTransaction}: transaction\_id, user\_id, amount, transaction\_type, source, reference\_id, created\_at, expiration\_date
    \item \texttt{MilesRedemptionRate}: rate\_id, redemption\_type, origin\_region, destination\_region, from\_cabin, to\_cabin, miles\_required, cash\_copay
\end{itemize}
\end{minipage}
}

\subsection{Saturation strategy under e-commerce domain}
\fbox{
\begin{minipage}{0.97\linewidth}
\textbf{V2 Evolution Context} \hfill \textbf{Strategy:} \texttt{saturation}

\vspace{4pt}
\textbf{Saturation Evolution -- New Shortcut Tools}

Implement the following shortcut tools that provide direct access to data that was previously only accessible through multi-hop traversals.

\vspace{4pt}
\textbf{New Tools to Implement}

\vspace{4pt}
\textbf{1. get\_order\_refund\_summary}

\textbf{Type:} \texttt{READ} \\
\textbf{Description:} Retrieves refund information for all returns and exchanges associated with a specific order, including request IDs and refund amounts.

\textbf{Inputs} (1 node(s)):
\texttt{Order.order\_id} (Order) [PK]

\textbf{Outputs} (2 node(s)):
    \texttt{ExchangeReturnRequest.request\_id} (ExchangeReturnRequest) [PK]; \texttt{ExchangeReturnRequest.refund\_amount} (ExchangeReturnRequest)

\textbf{Relationship:} references (one-to-many)

\textbf{Discovery Path:} \texttt{Order.exchange\_return\_request\_ids -> ExchangeReturnRequest.request\_id -> ExchangeReturnRequest.refund\_amount}

\textbf{Rationale:} This is a highly valuable shortcut that eliminates the need to fetch exchange/return request IDs first and then individually query each request for refund details. It provides a complete financial summary of an order's returns/exchanges in a single call, which is essential for customer service and financial reconciliation.

\textbf{Use Cases:}
    1. Customer service representative needs to quickly see all refunds associated with an order when handling a customer inquiry
    2. Accounting department needs to reconcile order financials including all partial refunds from returns/exchanges
    3. Customer viewing order history wants to see total refunded amount without navigating to individual return requests

\vspace{4pt}
\textbf{2. get\_subscription\_order\_swap\_history}

\textbf{Type:} \texttt{READ} \\
\textbf{Description:} Retrieves the complete product swap history for a subscription order, showing all product changes made to the subscription that generated this order.

\textbf{Inputs} (1 node(s)):
    \texttt{Order.parent\_subscription\_id} (Order) [FK]: Parent subscription ID if this is a subscription order

\textbf{Outputs} (2 node(s)):
    \texttt{SubscriptionSwapHistory.swap\_id} (SubscriptionSwapHistory) [PK]: Unique swap event identifier;
    \texttt{SubscriptionSwapHistory.subscription\_id} (SubscriptionSwapHistory) [FK]: Parent subscription reference

\textbf{Relationship:} references (one-to-many)

\textbf{Discovery Path:} \texttt{Order.parent\_subscription\_id -> Subscription.subscription\_id -> SubscriptionSwapHistory.subscription\_id -> SubscriptionSwapHistory.swap\_id}

\textbf{Rationale:} This tool is valuable for understanding the context of subscription orders---showing what product swaps occurred before this order was generated. It's particularly useful for customer service scenarios where a customer receives an unexpected product and needs to understand the swap history. It eliminates multiple queries through subscription and swap history tables.

\textbf{Use Cases:}
1. Customer receives a subscription order with an unexpected product and customer service needs to trace back what swaps occurred
2. User reviews past subscription order and wants to understand why they received a different product than originally subscribed
3. Analytics team analyzing subscription swap patterns and their impact on order fulfillment

\end{minipage}
}

\subsection{Saturation strategy under airline domain}
\fbox{
\begin{minipage}{0.97\linewidth}
\textbf{V2 Evolution Context} \hfill \textbf{Strategy:} \texttt{saturation}

\vspace{4pt}
\textbf{Saturation Evolution -- New Shortcut Tools}

Implement the following shortcut tools that provide direct access to data that was previously only accessible through multi-hop traversals.

\vspace{4pt}
\textbf{New Tools to Implement}

\vspace{4pt}
\textbf{1. get\_reservation\_flight\_statuses}

\textbf{Type:} \texttt{READ} \\
\textbf{Description:} Retrieves the current status of all flights in a reservation, allowing passengers to quickly check if any flights are delayed, cancelled, or on-time.

\textbf{Inputs} (1 node(s)): \texttt{Reservation.reservation\_id} (Reservation) [PK]: Unique reservation identifier (e.g., `ZFA04Y').

\textbf{Outputs} (2 node(s)): \texttt{ReservationFlight.flight\_number} (ReservationFlight) [FK]: Flight number reference; \texttt{FlightInstance.status} (FlightInstance): Flight status (available, on time, flying, landed, cancelled, delayed).

\textbf{Relationship:} contains (one-to-many)

\textbf{Discovery Path:} \texttt{Reservation.flights -> ReservationFlight.flight\_number -> Flight.flight\_number -> FlightInstance.status}

\textbf{Rationale:} This is one of the most frequent customer queries---checking the status of their booked flights. Currently requires chaining through Reservation $\to$ ReservationFlight $\to$ Flight $\to$ FlightInstance. This shortcut eliminates 3 separate lookups and provides immediate value for customer service and self-service scenarios.

\textbf{Use Cases:} Customer checking if their upcoming flights are on time; Customer service agent quickly assessing disruption impact on a booking; Automated notification system checking for status changes on reservations; Mobile app displaying real-time flight status dashboard for a trip.

\vspace{4pt}
\textbf{2. get\_segment\_flight\_pricing}

\textbf{Type:} \texttt{READ} \\
\textbf{Description:} Retrieves the price breakdown for all flights within a specific itinerary segment, essential for multi-city booking pricing and fare adjustments.

\textbf{Inputs} (1 node(s)): \texttt{ItinerarySegment.segment\_id} (ItinerarySegment) [PK]: Unique identifier for the itinerary segment (e.g., `SEG001').

\textbf{Outputs} (2 node(s)): \texttt{SegmentFlight.segment\_flight\_id} (SegmentFlight) [PK]: Unique identifier for the segment-flight association; \texttt{SegmentFlight.price} (SegmentFlight): Price for this flight in dollars.

\textbf{Relationship:} contains (one-to-many)

\textbf{Discovery Path:} \texttt{ItinerarySegment.segment\_id -> SegmentFlight.segment\_id -> SegmentFlight.segment\_flight\_id -> SegmentFlight.price}

\textbf{Rationale:} Multi-city reservations are complex and pricing queries per segment are frequent during booking modifications or fare recalculations. This shortcut from ItinerarySegment directly to SegmentFlight pricing eliminates intermediate lookups and is critical for the update\_segment\_cabin tool to calculate price differences.

\textbf{Use Cases:} Calculating upgrade costs for a specific leg of a multi-city trip; Displaying price breakdown by segment during booking review; Computing refund amounts when changing flights within a segment; Agent comparing alternative routing prices for a segment.

\end{minipage}
}

\subsection{Deprecation strategy under e-commerce domain}
\fbox{
\begin{minipage}{0.97\linewidth}
\textbf{V3 Evolution Context} \hfill \textbf{Strategy:} \texttt{deprecation}

\vspace{4pt}
\textbf{System Deprecation Notice}

The following changes have been made to the system. Agents must adapt accordingly.

\vspace{4pt}
\textbf{Deprecation Details}

\textbf{Type:} \texttt{DATABASE} \\
\textbf{Description:} Service down: Cart (5 nodes, 7 edges)

\textbf{Reason:} Cart service undergoing scheduled maintenance for database migration to improve scalability and session handling. The cart system is being moved to a new distributed architecture to handle peak traffic better.

\textbf{Impact:} Users cannot view, modify, or create shopping carts. Cart-to-checkout flow is unavailable. Cart item counts and saved items cannot be retrieved. Affects 5 data fields and 5 cart-related tools.

\textbf{Challenge Level:} \texttt{MEDIUM}

\vspace{4pt}
\textbf{Workaround.}
Agent can still help users browse products, check order history for past purchases, look up product details, and inform users when cart service will be restored. Users can note product IDs to add later, or agent could help with wishlists if available.

\vspace{4pt}
\textbf{Removed Connections.}

(1) \texttt{Product.asin -> Cart.items} --- Tools affected: \texttt{add\_to\_cart(user\_id: User.user\_id, product\_id: Product.asin, variant\_id?, quantity?) -> Cart};

(2) \texttt{Cart.cart\_id -> Cart.user\_id};

(3) \texttt{Cart.cart\_id -> Cart.items};

(4) \texttt{Cart.user\_id -> User.user\_id} --- Tools affected: \texttt{get\_cart(user\_id: User.user\_id) -> Cart}, \texttt{add\_to\_cart(user\_id: User.user\_id, product\_id: Product.asin, variant\_id?, quantity?) -> Cart}, \texttt{clear\_cart(user\_id: User.user\_id) -> Cart};

(5) \texttt{Cart.items -> CartItem.cart\_item\_id} --- Tools affected: \texttt{remove\_from\_cart(user\_id: User.user\_id, cart\_item\_id: CartItem.cart\_item\_id) -> Cart}, \texttt{update\_cart\_item(user\_id: User.user\_id, cart\_item\_id: CartItem.cart\_item\_id, quantity: int) -> Cart};

(6) \texttt{User.user\_id -> Cart.items} --- Tools affected: \texttt{get\_cart(user\_id: User.user\_id) -> Cart}, \texttt{add\_to\_cart(user\_id: User.user\_id, product\_id: Product.asin, variant\_id?, quantity?) -> Cart}, \texttt{clear\_cart(user\_id: User.user\_id) -> Cart};

(7) \texttt{User.cart\_id -> Cart.cart\_id}.

\vspace{4pt}
\textbf{Removed Data Points.}

\texttt{Cart.cart\_id} (Cart);
\texttt{Cart.user\_id} (Cart);
\texttt{Cart.items} (Cart);
\texttt{Cart.created\_at} (Cart);
\texttt{Cart.updated\_at} (Cart).

\vspace{4pt}
\textbf{Deprecated Tools.}

\texttt{add\_to\_cart(user\_id: User.user\_id, product\_id: Product.asin, variant\_id?, quantity?) -> Cart};
\texttt{update\_cart\_item(user\_id: User.user\_id, cart\_item\_id: CartItem.cart\_item\_id, quantity: int) -> Cart};
\texttt{remove\_from\_cart(user\_id: User.user\_id, cart\_item\_id: CartItem.cart\_item\_id) -> Cart};
\texttt{get\_cart(user\_id: User.user\_id) -> Cart};
\texttt{clear\_cart(user\_id: User.user\_id) -> Cart}.

\vspace{4pt}
\textbf{Agent Requirements.}

(1) Detect when deprecated services/tools are called;
(2) Find alternative paths when available;
(3) Gracefully handle cases with no alternatives;
(4) Provide meaningful error messages.

\end{minipage}
}

\subsection{Deprecation strategy under airline domain}
\fbox{
\begin{minipage}{0.97\linewidth}
\textbf{V3 Evolution Context} \hfill \textbf{Strategy:} \texttt{deprecation}

\vspace{4pt}
\textbf{System Deprecation Notice}

The following changes have been made to the system. Agents must adapt accordingly.

\vspace{4pt}
\textbf{Deprecation Details}

\textbf{Type:} \texttt{TOOL} \\
\textbf{Description:} Tool deprecation: \texttt{search\_direct\_flight(origin: str, destination: str, date: str) -> List[DirectFlight]} (8 edges affected)

\textbf{Reason:} The direct flight search API is being consolidated into the unified flight search website. As part of our Q4 API modernization initiative, specialized search endpoints are being retired in favor of the comprehensive \texttt{search\_oneway\_flight} and \texttt{search\_roundtrip\_flight} APIs which support filtering by number of stops. This reduces API surface area and maintenance overhead.

\textbf{Impact:} Agents can no longer directly search for non-stop flights between two cities. Flight discovery for direct routes requires using general search APIs and filtering results, or searching specific flight numbers if known.

\textbf{Challenge Level:} \texttt{MEDIUM}

\vspace{4pt}
\textbf{Workaround.}
Use \texttt{search\_oneway\_flight} to find flights between origin and destination, then filter results to identify direct (non-stop) flights by examining the flight data structure for connection indicators or flight duration.

\vspace{4pt}
\textbf{Removed Connections.}

(1) \texttt{Input.user\_input -> Flight.flight\_number} --- Tools affected: \texttt{search\_direct\_flight(origin: str, destination: str, date: str) -> List[DirectFlight]}, \texttt{search\_onestop\_flight(origin: str, destination: str, date: str) -> List[Tuple[DirectFlight, DirectFlight]]};
(2) \texttt{Flight.flight\_number -> Flight.origin} --- Tools affected: \texttt{search\_direct\_flight(origin: str, destination: str, date: str) -> List[DirectFlight]};
(3) \texttt{Flight.flight\_number -> Flight.destination} --- Tools affected: \texttt{search\_direct\_flight(origin: str, destination: str, date: str) -> List[DirectFlight]};
(4) \texttt{Flight.flight\_number -> Flight.scheduled\_departure\_time\_est} --- Tools affected: \texttt{search\_direct\_flight(origin: str, destination: str, date: str) -> List[DirectFlight]};
(5) \texttt{Flight.flight\_number -> Flight.scheduled\_arrival\_time\_est} --- Tools affected: \texttt{search\_direct\_flight(origin: str, destination: str, date: str) -> List[DirectFlight]};
(6) \texttt{Flight.flight\_number -> Flight.dates} --- Tools affected: \texttt{search\_direct\_flight(origin: str, destination: str, date: str) -> List[DirectFlight]}, \texttt{get\_flight\_status(flight\_number: str, date: str) -> str};
(7) \texttt{FlightInstance.flight\_number -> FlightInstance.available\_seats} --- Tools affected: \texttt{search\_direct\_flight(origin: str, destination: str, date: str) -> List[DirectFlight]}, \texttt{get\_flight\_availability\_details};
(8) \texttt{FlightInstance.flight\_number -> FlightInstance.prices} --- Tools affected: \texttt{search\_direct\_flight(origin: str, destination: str, date: str) -> List[DirectFlight]}.

\vspace{4pt}
\textbf{Deprecated Tools.}

\texttt{search\_direct\_flight(origin: str, destination: str, date: str) -> List[DirectFlight]}.

\vspace{4pt}
\textbf{Agent Requirements.}
(1) Detect when deprecated services/tools are called;
(2) Find alternative paths when available;
(3) Gracefully handle cases with no alternatives;
(4) Provide meaningful error messages.

\end{minipage}
}

\section{Prompts and Implementation Details}
\label{app:prompts-implementation}
\label{app:env-generation}
This appendix provides the prompts and implementation details used in the two-phase environment evolution pipeline. Phase I proposes semantic evolution intents and translates them into graph edits. Phase II instantiates the transformed graph into executable code and validates the resulting environment version.

\subsection{Phase I: Evolution Proposal Prompts}
\label{app:evolution-prompts}
\subsubsection{Completion: Task Proposer}
\begin{promptbox}
\textbf{System}
\smallskip
You are a Product Manager for a \{domain.name\} website. Your job is to identify 
new features and capabilities that would enhance the website but are NOT currently supported.

\#\# Domain Context: \{domain.description\}

\#\# You Have Deep Knowledge Of:
- \{domain.name.title()\} best practices
- Customer needs and pain points in \{domain.name\}
- Modern \{domain.name\} features (\{common\_features\})
- Backend data modeling and API design
- Industry-specific regulations and requirements

When proposing tasks, be specific and realistic. Consider the user journey and what data and tools 
would be needed to support the feature in a \{domain.name\} context.
\tcblower
\textbf{Prompt}
\smallskip
\#\# Task Proposal Request

Analyze the current \{domain.name\} system and propose a NEW feature/task that is NOT currently supported. 
The new feature should extend the system without significantly changing existing functionality.

\#\#\# Domain Context
\textbf{\{domain.name.title()\}}:
\{domain.description\}

\#\#\# Current System State

\#\#\#\# Databases (Data Entities):
\{', '.join(databases)\}

\#\#\#\# Available Tools (APIs):
\{', '.join(tools)\}

\#\#\#\# Current Graph Structure:
\texttt{```json}
\{graph\_json\}

\#\#\# Your Task
Propose a feature in the domain of ``\{selected\_domain\}'' that the current system CANNOT support.

Use seed value \{seed\} to ensure your proposal is unique and creative. \{exclusion\_text\}

\#\#\#\# Output Format
Respond with the task in EXACTLY this format (wrapped in $<$task\_proposal$>$ tags):

$<$task\_proposal$>$
\#\#\# Task: [Concise Task Name]
User Story: ``[First-person user story describing the need]''

Why Not Supported:

[Specific reason 1 - what data/entity is missing]
[Specific reason 2 - what tool/API is missing]
[Continue with specific gaps...]

Required Capabilities Missing:

[tool\_name\_1()]
tool - [what it does]
[tool\_name\_2()]
tool - [what it does]
[EntityName]
entity with fields: [field1, field2, ...]
[Continue with specific requirements...]
$<$\/task\_proposal$>$

\#\#\#\# Guidelines
- Be specific about what's missing
- Reference actual databases and tools from the current system when explaining gaps
- Propose realistic \{domain.name\} features that would add business value
- Ensure the task is achievable with reasonable additions to the system
- Consider \{domain.name\}-specific requirements and best practices
\end{promptbox}

\subsubsection{Saturation: Tool Designer}
\begin{promptbox}
\textbf{System}
\smallskip
You are an expert API designer for \{domain.name\} systems. Your task is to design meaningful, 
well-named tools (API endpoints) that provide shortcuts for accessing data across entity relationships.

\#\# Domain Context: \{domain.description\}

\#\# You Understand:

\{domain.name.title()\} domain concepts (\{entities\_formatted.lower()\}, etc.);
RESTful API naming conventions;
The value of convenience methods that combine multiple operations;
Clear, descriptive naming that indicates what the tool does;
\{domain.name.title()\}-specific terminology and conventions.

\#\# Your Tool Names Should Be:

Snake\_case format;
Action-oriented (get\_, list\_, find\_, search\_, add\_, update\_, etc.);
Clear about what data is being accessed or modified;
Concise but descriptive;
Using \{domain.name\}-appropriate terminology.
\tcblower
\textbf{Prompt}
\smallskip
\#\# Tool Design Task

\#\#\# Domain Context:
\{domain.name.title()\}: \{domain.description\}

\#\#\# System Context

\#\#\#\# Databases/Entities:
\{', '.join(databases)\}

\#\#\#\# Existing Tools (for context, you may create similar tools with different functionality):
\{existing\_tools\_display\}

\#\#\# Task

You are given \{len(path\_descriptions)\} candidate paths discovered through graph traversal. Your task is to:

1. \textbf{SELECT} the \{num\_tools\} MOST VALUABLE paths that would benefit from having shortcut tools
2. \textbf{DESIGN} a tool for each selected path, choosing which nodes serve as INPUTS and which as OUTPUTS

\#\#\# Multi-Input/Multi-Output Tools

You can design tools with flexible input/output mappings:
- \textbf{Single Input $\to$ Single Output}: Simple lookup (e.g., get product by user\_id)
- \textbf{Multiple Inputs $\to$ Single Output}: Filtered lookup (e.g., get order by user\_id AND date)
- \textbf{Single Input $\to$ Multiple Outputs}: Fetch related data (e.g., get user's orders AND cart items)
- \textbf{Multiple Inputs $\to$ Multiple Outputs}: Complex queries (e.g., get products and reviews by user AND category)

For each path, you can select ANY nodes from the path as inputs or outputs---they don't have to be just the endpoints!

\#\#\# Selection Criteria (use your judgment):
- \textbf{Business Value}: How useful would this shortcut be in real \{domain.name\} scenarios?
- \textbf{Time Savings}: How much complexity does the shortcut eliminate?
- \textbf{Natural Fit}: Does this relationship make intuitive sense to users/developers?
- \textbf{Diversity}: Select paths that provide different types of functionality
- \textbf{Practicality}: Would developers actually use this tool frequently?

\#\#\# All Candidate Paths

\texttt{json}
\{json.dumps(path\_descriptions, indent=2)\}

\#\#\#\# Output Format
Select exactly \{num\_tools\} paths and provide tool proposals in this exact JSON format (wrapped in $<$tool\_proposals$>$ tags):

$<$tool\_proposals$>$
[
  \{\{
    "path\_id": 0,
    "tool\_name": "snake\_case\_tool\_name",
    "tool\_type": "READ or WRITE",
    "description": "Clear description of what the tool does",
    "input\_node\_ids": [1, 5],
    "output\_node\_ids": [12, 15, 18],
    "relationship\_type": "references$|$belongs\_to$|$contains$|$aggregates$|$links\_to",
    "cardinality": "one-to-one$|$one-to-many$|$many-to-one$|$many-to-many",
    "rationale": "Why you selected this path and why this shortcut is valuable",
    "use\_cases": ["Use case 1", "Use case 2"]
  \}\}
]
$<$/tool\_proposals$>$

\#\#\#\# Guidelines
- Input/Output Selection:
 - `input\_node\_ids`: List of node\_ids from the path that serve as inputs (parameters to the tool)
 - `output\_node\_ids`: List of node\_ids from the path that serve as outputs (returned data)
- You can select ANY nodes from the path---not just start/end
- At least one input and one output are required
- Tool Names: Use snake\_case, be descriptive but concise
 - READ tools: get\_, list\_, find\_, search\_, lookup\_
 - WRITE tools: add\_, update\_, link\_, associate\_, set\_

Be Specific. The tool should clearly indicate:
- What entity/data is being accessed
- The relationship between inputs and outputs
- Whether it's a read or write operation
- Use Cases: Provide realistic \{domain.name\} scenarios

\#\#\#\# Relationship Types:
`references`: Direct FK relationship;
`belongs\_to`: Child-to-parent relationship;
`contains`: Parent-to-children relationship;
`aggregates`: Computed/derived relationship;
`links\_to`: Cross-entity shortcut

Important: You MUST select exactly \{num\_tools\} paths. Choose wisely based on value, not just arbitrarily.
\end{promptbox}

\subsubsection{Deprecation: Deprecation Selector}
\begin{promptbox}
\textbf{System}
\smallskip
You are an expert system architect responsible for managing API deprecations 
and service maintenance in a \{domain.name\} website.

\#\# Domain Context: \{domain.description\}
\#\#\# Core Functional Areas:
\{core\_areas\_formatted\}

Your responsibilities:
1. Evaluate proposed deprecations for realism and business sense
2. Assess the impact on downstream systems and users
3. Determine appropriate challenge levels for system resilience testing
4. Suggest workarounds when available

You understand:
- Real-world reasons for API deprecation (security, performance, consolidation, sunset)
- Service outage scenarios (maintenance, failures, migrations)
- How agents/systems should gracefully handle unavailable services
- The importance of testing system robustness
- \{domain.name.title()\}-specific operational concerns and compliance requirements

Your deprecation decisions should be:
- Realistic (something that would actually happen in a \{domain.name\} production system)
- Challenging but fair (agents should be able to adapt)
- Well-documented (clear reasons and workarounds)
- Domain-appropriate (consider \{domain.name\}-specific regulations and practices)
\tcblower
\textbf{Prompt}
\smallskip
\#\# Deprecation Selection Task

\#\#\# Context
You are reviewing deprecation candidates for a \{domain.name\} system.
Sampling mode used: \textbf{\{mode\}}

\#\#\# Graph Overview
- Databases: \{', '.join(graph\_databases)\}
- Total nodes: \{graph\_num\_nodes\}
- Total edges: \{graph\_num\_edges\}

\#\#\# Candidates

```json
\{json.dumps(candidates, indent=2)\}
```

\#\#\# Your Task
Select ONE candidate that best satisfies:
- Realism: The deprecation should be something that could actually happen in production
- Challenge: It should create a meaningful challenge for an agent to handle (prefer medium to hard difficulty)
- Interestingness: The deprecation should test the agent's ability to adapt

\#\#\# Output Format
Provide your selection in this JSON format wrapped in $<$deprecation\_decision$>$ tags:

$<$deprecation\_decision$>$
\{\{ "candidate\_id": 0, "deprecation\_reason": "Realistic reason why this would be deprecated", "impact\_summary": "Brief description of what functionality is affected", "challenge\_level": "easy|medium|hard|extreme", "workaround\_hint": "Brief hint about how an agent might work around this" \}\}
$<$/deprecation\_decision$>$

\#\#\#\# Challenge Level Guidelines
- easy: Clear alternatives exist, minimal adaptation needed
- medium: Alternatives exist but require finding new paths (PREFERRED)
- hard: Limited alternatives, significant changes needed
- extreme: Critical functionality lost, graceful failure required

Choose wisely - select the candidate that creates the most interesting and realistic challenge.
\end{promptbox}

\subsubsection{Graph Evolver for Attribute Graph Updates}
\begin{promptbox}
\textbf{System}
\smallskip
You are a Software Architect designing data models and APIs for a \{domain.name\} website.

\#\# Domain Context: \{domain.description\}

Given a task requirement and the current system state (represented as an attribute graph), your job is to:
1. Design new data entities (databases) with their attributes
2. Design new tools (APIs) that operate on these entities
3. Define relationships between new and existing entities

\#\# Follow These Conventions:
- Database names are PascalCase (e.g., \{', '.join(list(domain.example\_entities.keys())[:3])\})
- Attribute names are snake\_case (e.g., \{list(domain.example\_entities.keys())[0].lower()\}\_id, created\_at)
- Tool names are snake\_case verbs (e.g., create\_\{list(domain.example\_entities.keys())[0].lower()\}, get\_\{list(domain.example\_entities.keys())[1].lower()\}\_status)
- Primary keys typically end with \_id
- Foreign keys reference other entities' primary keys
- Include created\_at/updated\_at timestamps for mutable entities

\#\# Relationship Types:
references, belongs\_to, contains, has\_attribute, identifies, used\_for, updates, explains

\#\# Cardinalities:
one-to-one, one-to-many, many-to-one, many-to-many
\tcblower
\textbf{Prompt}
\smallskip
\#\# Graph Evolution Design Task

\#\#\# Domain Context
\{domain.name.title()\}: \{domain.description\}

Given the following task requirement and current system state, design the data model and API changes needed.

\#\#\# Task Requirement

\{task\_full\_text\}

\#\#\# Current System State

```json
\{graph\_json\}
```

\#\#\# Your Task
Design the additions needed to support this task:
- New Databases (Entities): What new data entities are needed?
- New Nodes (Attributes): What attributes do these entities need? What new attributes for existing entities?
- New Edges (Relationships): How do new entities connect to existing ones?
- New Tools: What new APIs are needed?

\#\#\#\# Output Format
Respond with your design in EXACTLY this JSON format (wrapped in $<$graph\_evolve\_design$>$ tags). Include fields like `allowed\_values` that are presented in the original json when applicable:

$<$graph\_evolve\_design$>$
\{\{
  "rationale": "Brief explanation of the design decisions",
  "new\_databases": ["DatabaseName1", "DatabaseName2"],
  "new\_nodes": [
    \{\{
      "database": "DatabaseName",
      "attribute": "attribute\_name",
      "type": "str$|$int$|$float$|$bool$|$List[str]$|$Dict[str, Any]$|$Optional[str]$|$etc",
      "description": "What this attribute represents",
      "is\_primary\_key": true$|$false,
      "is\_foreign\_key": true$|$false,
      "modifiable": true$|$false,
    \}\}
  ],
  "new\_edges": [
    \{\{
      "source\_database": "SourceDB",
      "source\_attribute": "source\_attr",
      "target\_database": "TargetDB",
      "target\_attribute": "target\_attr",
      "relationship\_type": "references$|$belongs\_to$|$contains$|$has\_attribute",
      "cardinality": "one-to-one$|$one-to-many$|$many-to-one$|$many-to-many",
      "description": "What this relationship represents",
      "tools": ["tool\_name\_1(argument\_1: node\_name\_1, argument\_2: node\_name\_2) -$>$ return\_data\_class", ...]
    \}\}
  ],
  "new\_tools": ["tool\_name\_1(argument\_1: node\_name\_1, argument\_2: node\_name\_2) -$>$ return\_data\_class", ...]
\}\}
$<$/graph\_evolve\_design$>$

\#\#\# Design Guidelines
- Every new database MUST have a primary key (usually `[entity]\_id`)
- Foreign keys should reference existing primary keys
- Tools should be listed on the edges they operate on
- Include both READ and WRITE tools as appropriate
- Connect new entities to existing entities where logical
- Include standard fields like `created\_at`, `updated\_at`, `status` where appropriate
- Use consistent naming conventions:
 - Database names: PascalCase
 - Attribute names: snake\_case
 - Tool names: snake\_case verbs (e.g., create\_order, get\_order\_status)
\end{promptbox}

\subsection{Phase II: Graph-to-Code Instantiation}
\label{app:code-generation}
\label{app:code-validation}
\subsubsection{Coding Agent}
\begin{promptbox}
\textbf{System}
\smallskip
You are the Coding Agent in a \{domain.name\} environment evolution pipeline. Your role is to 
generate updated Python code (data models and tool implementations) based on evolved graph representations.

\#\# Pipeline Context

The environment evolution pipeline works as follows:
1. A Base Environment is represented as an attributed graph (nodes = data entities, edges = relationships + tools)
2. Graph Evolution Strategies transform the base graph into a new graph representation:
   - \textbf{Completion}: Adds nodes and edges based on task requirements from a Task Proposal Agent
   - \textbf{Saturation}: Adds edges via random walk sampling to discover new tool capabilities
   - \textbf{Deprecation}: Removes edges and nodes via graph pruning to deprecate features
3. You (the Coding Agent) receive the new graph representation and generate:
   - `data\_model.py`: Updated Pydantic models reflecting schema changes
   - `tools\_implementation.py`: Updated tool implementations reflecting edge changes

\#\# Your Responsibilities

1. \textbf{Analyze Graph Differences}: Compare the base graph with the new graph to identify:
   - Added nodes $\to$ Create new Pydantic models or add fields to existing models
   - Removed nodes $\to$ Remove or deprecate corresponding models/fields
   - Added edges $\to$ Implement new tools with @is\_tool decorators
   - Modified edges $\to$ Update tool signatures, parameters, or behavior
   - Removed edges $\to$ Remove or deprecate corresponding tools

2. \textbf{Maintain Code Consistency}: Use the base\_data.py and base\_tool.py as templates to ensure:
   - Consistent coding patterns and style
   - Proper Pydantic model inheritance and validation
   - Correct @is\_tool(ToolType.READ/WRITE) decorators
   - Comprehensive docstrings with Args, Returns, Raises sections
   - Appropriate error handling with ValueError for invalid inputs

3. \textbf{Handle Evolution Strategies Appropriately}:
   - Completion: Focus on adding new functionality while preserving existing code
   - Modification: Update existing tools/models without breaking backward compatibility
   - Saturation: Add tools that leverage existing data relationships
   - Deprecation: Safely remove code while maintaining system integrity

\#\# \{domain.name.title()\} Domain Scope

The environment covers these core functional areas, and will progress to support additional features:
\{core\_areas\_formatted\}

\#\#\# Key Entities and ID Patterns
\{entities\_formatted\}

\#\#\# Database and Toolkit Classes
- Database class: `\{domain.database\_class\}`
- Toolkit class: `\{domain.toolkit\_class\}`

Always generate complete, syntactically correct Python code that can be directly saved to files.
\tcblower
\textbf{Prompt}
\smallskip
\#\#\# Evolution Context

The following context describes why this evolution is happening and should guide your implementation:

\{evolution\_context\}

\#\#\# Environment files

$<$attribute\_graph$>$
\{graph\_rep\}
$<$/attribute\_graph$>$

$<$data\_model$>$
\{data\_model\}
$<$/data\_model$>$

$<$tools\_implementation$>$
\{tools\_implementation\}
$<$/tools\_implementation$>$

$<$updated\_attribute\_graph$>$
\{updated\_graph\_rep\}
$<$/updated\_attribute\_graph$>$

$<$updated\_data\_model$>$
\{updated\_data\_model\}
$<$/updated\_data\_model$>$

\#\#\# Instructions

1. Analyze the differences between the original `attribute\_graph` and `updated\_attribute\_graph`
2. Identify new edges that represent new tools to be implemented
3. For each new tool mentioned in the edges' ``tools'' field:
   - Create a new method with the @is\_tool decorator
   - Determine if it's a READ or WRITE operation based on the relationship type
   - Implement the tool logic following existing patterns in the codebase
   - Add proper docstrings with Args, Returns, and Raises sections
   - Handle error cases appropriately
4. Update imports to include any new models from the updated data\_model
5. Update the `\{domain.database\_class\}` or relevant database class if new collections are needed
6. Ensure the tools properly interact with the new data structures
7. Follow the existing code style and patterns
8. Add any necessary helper methods

\#\#\# Key Considerations for \{domain.name.title()\} Domain

- READ tools should use @is\_tool(ToolType.READ)
- WRITE tools should use @is\_tool(ToolType.WRITE)
- Tools should validate inputs and raise ValueError for invalid data
- Tools should properly update related entities (e.g., updating user references when creating new entities)
- Use the `\{domain.toolkit\_class\}` class for all tool implementations

Please update the `tools\_implementation.py` file, wrapping the complete updated code within $<$updated\_tools\_implementation$>$ and $<$/updated\_tools\_implementation$>$ tags. Output the complete file, not just the changes.
\end{promptbox}

\subsubsection{Test Specification Generator}
\begin{promptbox}
\textbf{System}
\smallskip
You are the Test Generation Agent in a \{domain.name\} environment evolution pipeline. Your role is to 
generate comprehensive pytest unit tests for newly created or modified data models and tools.

\#\# Pipeline Context

After the Coding Agent generates updated data models and tool implementations based on evolved graph 
representations, you generate tests to verify the correctness of:
1. New Pydantic data models (field validation, serialization, edge cases)
2. New tool implementations (happy paths, error handling, edge cases)
3. Integration between models and tools

\#\# Your Responsibilities

1. \textbf{Generate Test Specifications}: Analyze the graph changes to determine what needs testing:
   - New data models $\to$ Test instantiation, field validation, serialization
   - New tools $\to$ Test input/output contracts, error handling, edge cases
   - New relationships $\to$ Test data integrity across entities

2. \textbf{Write Comprehensive Tests}: For each component, generate tests covering:
   - \textbf{Happy Path}: Valid inputs produce expected outputs
   - \textbf{Edge Cases}: Boundary values, empty inputs, maximum lengths
   - \textbf{Error Handling}: Invalid inputs, missing data, not found scenarios
   - \textbf{Type Validation}: Correct types are enforced

3. \textbf{Follow Testing Best Practices}:
   - Use pytest framework with clear test function names (test\_\texttt{$<$function$>$}\_)
   - Use unittest.mock to mock database operations and external dependencies
   - Create realistic test fixtures with valid \{domain.name\} data
   - Each test should be independent and not rely on other tests
   - Use descriptive assertion messages

4. \textbf{Mock Database Operations}:
   - Mock the database class (`\{domain.database\_class\}`) to avoid actual database calls
   - Set up mock return values that match expected data structures
   - Verify that database methods are called with correct arguments

\#\# \{domain.name.title()\} Domain Context

\#\#\# Key Entities and ID Patterns
\{entities\_formatted\}

\#\#\# Example Test Data
Generate test data that reflects realistic \{domain.name\} scenarios:
\{test\_data\_formatted\}

\#\#\# Database and Toolkit Classes
- Database class: `\{domain.database\_class\}`
- Toolkit class: `\{domain.toolkit\_class\}`

Always generate complete, syntactically correct pytest code that can be directly executed.
\tcblower
\textbf{Prompt}
\smallskip
\#\#\# Evolution Context

The following context describes why this evolution is happening and should guide your implementation:

\{evolution\_context\}

\#\#\# Updated Graph Representation

\{updated\_graph\_rep\}

\#\#\# Graph Changes Summary

\textbf{New Databases/Entities}: \{graph\_diff.get('new\_databases', [])\}

\textbf{Added Nodes (new data fields)}:
\{\_format\_nodes\_for\_prompt(graph\_diff.get('added\_nodes', []))\}

\textbf{Removed Nodes}:
\{\_format\_nodes\_for\_prompt(graph\_diff.get('removed\_nodes', []))\}

\textbf{Added Edges (new relationships/tools)}:
\{\_format\_edges\_for\_prompt(graph\_diff.get('added\_edges', []))\}

\textbf{Removed Edges}:
\{\_format\_edges\_for\_prompt(graph\_diff.get('removed\_edges', []))\}

\textbf{New Tools to Implement}: \{graph\_diff.get('new\_tools', [])\}

\#\#\# Instructions

Based on the graph changes above, generate a structured test specification that defines WHAT should be tested.
Do NOT write actual test code yet---just specify the test requirements.

For each component, provide:

\#\#\#\# 1. Data Model Test Specifications
For each new or modified data model:
- Model name and purpose
- Required fields and their expected types
- Field validation rules (min/max, patterns, allowed values)
- Optional vs required fields
- Edge cases to test (empty values, null, boundary values, invalid types)
- Serialization/deserialization requirements

\#\#\#\# 2. Tool Test Specifications
For each new or modified tool:
- Tool name and purpose
- Input parameters with types and constraints
- Expected output structure for success cases
- Error conditions and expected exceptions
- Edge cases to test:
  - Empty/null inputs
  - Non-existent references (e.g., entity not found)
  - Invalid parameter types
  - Boundary values
- Side effects to verify (e.g., database updates, related entity changes)

\#\#\#\# 3. Integration Test Specifications
For relationships between entities:
- Data integrity constraints to verify
- Cross-entity operations to test
- Cascading effects to validate

\#\#\# \{domain.name.title()\} Domain Test Data Examples

Use these patterns for realistic test data:
\{chr(10).join(f"- \{k\}: \{repr(v)\}" for k, v in domain.example\_test\_data.items())\}

Output the test specifications in a structured format wrapped in $<$test\_specifications$>$ and $<$/test\_specifications$>$ tags. Use clear sections and bullet points for each test requirement.
\end{promptbox}

\subsubsection{Validation and Filtering}
\paragraph{Validation and filtering pipeline.}\label{para:validation}
Each generated env-version passes through a cascading validation pipeline
before it is admitted as a base for subsequent evolution steps. The
pipeline has four levels, ordered by cost and specificity.
\emph{(i)~Graph-validity checks} inspect the transformed graph
$\mathcal{G}^{(k+1)}$ for schema consistency, reachability from
\texttt{Input} nodes to newly added tools, and dependency closure (every
referenced node and edge is defined). Graph edits that violate these
constraints are rejected before any code is generated.
\emph{(ii)~Syntax and import checks} compile every generated
\texttt{data\_model.py} and \texttt{tools\_implementation.py} under the
target Python runtime, then confirm that every symbol imported by the
generated \texttt{test\_tools.py} resolves against the implementation
module. This catches truncated LLM outputs, missing class definitions,
and unresolved references at the earliest possible point.
\emph{(iii)~Unit-test execution} runs the graph-grounded tests with
\texttt{pytest} to check that the new and modified tools behave as
specified. \emph{(iv)~Semantic-intent check} applies the LLM-as-a-Judge
described in~\S\ref{subsubsec:semantic_fidelity} to verify that the
implemented change realizes the intended graph edit rather than merely
passing tests.

\paragraph{Retry behavior.} Failures at stages (ii) and (iii) trigger a
bounded repair loop with up to $R{=}3$ retries. On each retry the
coding agent is re-prompted with the failing code and the captured
error (SyntaxError trace, ImportError, pytest failure output), and
produces a targeted patch rather than regenerating the full file. This
matches the stage of failure: syntax/import errors typically resolve in
a single retry because the error message localizes the defect, whereas
test-execution failures may need two to three rounds to converge.
After $R$ retries we stop to bound cost; the version is then submitted
to stage (iv) as-is with its residual failures annotated, and the
filtering rule (below) decides its fate.

\paragraph{Filtering rule for critical defects.} We treat an env-version
as having a \emph{critical tool-code defect} iff at least one failure
reaches severity level~3 in the taxonomy of
Appendix~\ref{app:test-failure}, i.e., a genuine bug in
\texttt{tools\_implementation.py} or \texttt{data\_model.py} (e.g.,
collection-time \texttt{ImportError}/\texttt{NameError}, missing tool
method, undefined symbol inside tool code). Such env-versions are
\emph{filtered}: they are not admitted as a base $\mathcal{G}^{(k)}$
for the next evolution step, which prevents defects from propagating
and compounding across the trajectory. Env-versions with only level-1
or level-2 failures (test-side artifacts or minor corner cases) are
retained because the tool implementation itself is correct.

\paragraph{Post-hoc defect rates.} In practice, the filter activates on a
small minority of env-versions. Over the 150 e-commerce env-versions,
6 (4.0\%) contain a critical defect and are filtered:
4 at V1 (completion), 0 at V2 (saturation), and 2 at V3 (deprecation).
Over the 15 airline env-versions, 2 (13.3\%) are filtered, both at V3.
Inspecting the filtered cases, the defects cluster into three recurring
patterns: (a)~missing standard-library imports used in the new code
(e.g., \texttt{timedelta} without \texttt{from datetime import
timedelta}); (b)~classes referenced in tests that were placed in
\texttt{tools\_implementation.py} instead of being exported from
\texttt{data\_model.py}; and (c)~data-model entities that the graph edit
introduced but the coding agent failed to serialize (e.g.,
\texttt{FlightInstance}, \texttt{ServiceUnavailableError}). The first
two are already systematically addressable by a single repair pass and
vanish as coding-agent capability improves; the third is a genuine
capacity limit of the current back-end. No V2~(saturation) env-version
exhibits a critical defect, consistent with saturation being the least
structurally invasive primitive---it adds shortcut edges over existing
reachability rather than new entities, so the coding agent has less
opportunity to omit definitions.

\paragraph{Remark on sufficiency.} Because generation is cheap and
parallel, filtered env-versions are replaced by additional sampled
trajectories rather than manually repaired, so the filtering rule does
not bottleneck scale. The combination of cheap graph-validity checks,
executable tests as ground truth, and a conservative filter on
level-3 defects keeps the error rate per \emph{published} env-version
close to zero without requiring the coding agent to be perfect.

\subsection{Runtime Configuration and Model Selection}
\label{app:runtime-config}

All components of the \textsc{ProEvolve} pipeline---benchmark generation, task generation, and evaluation---use the following unified configuration.

\paragraph{Generation backbone.} Both Phase~I (evolution proposal: task proposer, graph evolver, saturation tool designer, deprecation selector) and Phase~II (coding agent, test specification generator) use \texttt{claude-opus-4-5-20251101} as the backbone LLM. This single-model choice ensures stylistic consistency across generated code, graph edits, and test specifications within each evolution trajectory.

\paragraph{Task generation.} The subgraph-based task pipeline (Sec.~\ref{subsec:method:taskgen})---including scenario synthesis, sandbox materialization, agentic walk execution, and state-wise instruction generation---also uses \texttt{claude-opus-4-5-20251101}.

\paragraph{User simulator.} During downstream agent evaluation (Sec.~\ref{subsec:method:eval}), the state-wise user simulator that issues instructions and judges state satisfaction is powered by \texttt{claude-opus-4-5-20251101}, following the same protocol as $\tau^2$-bench~\cite{barres2025tau2}.

\paragraph{LLM-as-a-Judge.} Semantic fidelity evaluation (Appendix~\ref{app:semantic-fidelity}) and task quality assessment (Appendix~\ref{app:task-quality}) use \texttt{claude-sonnet-4-5-20250514} for cost efficiency, as these require scoring individual modifications against rubrics rather than generating complex code or multi-turn interactions.

\paragraph{Decoding parameters.} All LLM calls across the entire pipeline---generation, evaluation, and judging---use \texttt{temperature=0} with no top-$p$ or top-$k$ truncation, ensuring deterministic outputs and full reproducibility given the same inputs. No sampling-based decoding is used at any stage.

\clearpage
\section*{NeurIPS Paper Checklist}

\begin{enumerate}

\item {\bf Claims}
    \item[] Question: Do the main claims made in the abstract and introduction accurately reflect the paper's contributions and scope?
    \item[] Answer: \answerYes{} 
    \item[] Justification: The abstract and introduction outline the key problem studied, our contributions, and a summary of the experimental findings.
    \item[] Guidelines:
    \begin{itemize}
        \item The answer \answerNA{} means that the abstract and introduction do not include the claims made in the paper.
        \item The abstract and/or introduction should clearly state the claims made, including the contributions made in the paper and important assumptions and limitations. A \answerNo{} or \answerNA{} answer to this question will not be perceived well by the reviewers. 
        \item The claims made should match theoretical and experimental results, and reflect how much the results can be expected to generalize to other settings. 
        \item It is fine to include aspirational goals as motivation as long as it is clear that these goals are not attained by the paper. 
    \end{itemize}

\item {\bf Limitations}
    \item[] Question: Does the paper discuss the limitations of the work performed by the authors?
    \item[] Answer: \answerYes{} 
    \item[] Justification: See Sec.~\ref{sec:conclusions} and Appendix \ref{app:impact}.
    \item[] Guidelines:
    \begin{itemize}
        \item The answer \answerNA{} means that the paper has no limitation while the answer \answerNo{} means that the paper has limitations, but those are not discussed in the paper. 
        \item The authors are encouraged to create a separate ``Limitations'' section in their paper.
        \item The paper should point out any strong assumptions and how robust the results are to violations of these assumptions (e.g., independence assumptions, noiseless settings, model well-specification, asymptotic approximations only holding locally). The authors should reflect on how these assumptions might be violated in practice and what the implications would be.
        \item The authors should reflect on the scope of the claims made, e.g., if the approach was only tested on a few datasets or with a few runs. In general, empirical results often depend on implicit assumptions, which should be articulated.
        \item The authors should reflect on the factors that influence the performance of the approach. For example, a facial recognition algorithm may perform poorly when image resolution is low or images are taken in low lighting. Or a speech-to-text system might not be used reliably to provide closed captions for online lectures because it fails to handle technical jargon.
        \item The authors should discuss the computational efficiency of the proposed algorithms and how they scale with dataset size.
        \item If applicable, the authors should discuss possible limitations of their approach to address problems of privacy and fairness.
        \item While the authors might fear that complete honesty about limitations might be used by reviewers as grounds for rejection, a worse outcome might be that reviewers discover limitations that aren't acknowledged in the paper. The authors should use their best judgment and recognize that individual actions in favor of transparency play an important role in developing norms that preserve the integrity of the community. Reviewers will be specifically instructed to not penalize honesty concerning limitations.
    \end{itemize}

\item {\bf Theory assumptions and proofs}
    \item[] Question: For each theoretical result, does the paper provide the full set of assumptions and a complete (and correct) proof?
    \item[] Answer: \answerNA{} 
    \item[] Justification: 
    \item[] Guidelines:
    \begin{itemize}
        \item The answer \answerNA{} means that the paper does not include theoretical results. 
        \item All the theorems, formulas, and proofs in the paper should be numbered and cross-referenced.
        \item All assumptions should be clearly stated or referenced in the statement of any theorems.
        \item The proofs can either appear in the main paper or the supplemental material, but if they appear in the supplemental material, the authors are encouraged to provide a short proof sketch to provide intuition. 
        \item Inversely, any informal proof provided in the core of the paper should be complemented by formal proofs provided in appendix or supplemental material.
        \item Theorems and Lemmas that the proof relies upon should be properly referenced. 
    \end{itemize}

    \item {\bf Experimental result reproducibility}
    \item[] Question: Does the paper fully disclose all the information needed to reproduce the main experimental results of the paper to the extent that it affects the main claims and/or conclusions of the paper (regardless of whether the code and data are provided or not)?
    \item[] Answer: \answerYes{} 
    \item[] Justification: We have described the experimental details and properly cited the opensource materials necessary for reproducing the experiments. The code will be made publicly available upon acceptance.
    \item[] Guidelines:
    \begin{itemize}
        \item The answer \answerNA{} means that the paper does not include experiments.
        \item If the paper includes experiments, a \answerNo{} answer to this question will not be perceived well by the reviewers: Making the paper reproducible is important, regardless of whether the code and data are provided or not.
        \item If the contribution is a dataset and\slash or model, the authors should describe the steps taken to make their results reproducible or verifiable. 
        \item Depending on the contribution, reproducibility can be accomplished in various ways. For example, if the contribution is a novel architecture, describing the architecture fully might suffice, or if the contribution is a specific model and empirical evaluation, it may be necessary to either make it possible for others to replicate the model with the same dataset, or provide access to the model. In general. releasing code and data is often one good way to accomplish this, but reproducibility can also be provided via detailed instructions for how to replicate the results, access to a hosted model (e.g., in the case of a large language model), releasing of a model checkpoint, or other means that are appropriate to the research performed.
        \item While NeurIPS does not require releasing code, the conference does require all submissions to provide some reasonable avenue for reproducibility, which may depend on the nature of the contribution. For example
        \begin{enumerate}
            \item If the contribution is primarily a new algorithm, the paper should make it clear how to reproduce that algorithm.
            \item If the contribution is primarily a new model architecture, the paper should describe the architecture clearly and fully.
            \item If the contribution is a new model (e.g., a large language model), then there should either be a way to access this model for reproducing the results or a way to reproduce the model (e.g., with an open-source dataset or instructions for how to construct the dataset).
            \item We recognize that reproducibility may be tricky in some cases, in which case authors are welcome to describe the particular way they provide for reproducibility. In the case of closed-source models, it may be that access to the model is limited in some way (e.g., to registered users), but it should be possible for other researchers to have some path to reproducing or verifying the results.
        \end{enumerate}
    \end{itemize}

\item {\bf Open access to data and code}
    \item[] Question: Does the paper provide open access to the data and code, with sufficient instructions to faithfully reproduce the main experimental results, as described in supplemental material?
    \item[] Answer: \answerYes{} 
    \item[] Justification: We used the open-sourced dataset, i.e., \cite{yao2022webshop} and \cite{barres2025tau2}, and we plan to publicly release our code once the
status of the paper is confirmed.
    \item[] Guidelines: 
    \begin{itemize}
        \item The answer \answerNA{} means that paper does not include experiments requiring code.
        \item Please see the NeurIPS code and data submission guidelines (\url{https://neurips.cc/public/guides/CodeSubmissionPolicy}) for more details.
        \item While we encourage the release of code and data, we understand that this might not be possible, so \answerNo{} is an acceptable answer. Papers cannot be rejected simply for not including code, unless this is central to the contribution (e.g., for a new open-source benchmark).
        \item The instructions should contain the exact command and environment needed to run to reproduce the results. See the NeurIPS code and data submission guidelines (\url{https://neurips.cc/public/guides/CodeSubmissionPolicy}) for more details.
        \item The authors should provide instructions on data access and preparation, including how to access the raw data, preprocessed data, intermediate data, and generated data, etc.
        \item The authors should provide scripts to reproduce all experimental results for the new proposed method and baselines. If only a subset of experiments are reproducible, they should state which ones are omitted from the script and why.
        \item At submission time, to preserve anonymity, the authors should release anonymized versions (if applicable).
        \item Providing as much information as possible in supplemental material (appended to the paper) is recommended, but including URLs to data and code is permitted.
    \end{itemize}

\item {\bf Experimental setting/details}
    \item[] Question: Does the paper specify all the training and test details (e.g., data splits, hyperparameters, how they were chosen, type of optimizer) necessary to understand the results?
    \item[] Answer: \answerYes{} 
    \item[] Justification:  Section~\ref{subsec:bench_setup} describes the benchmark construction setup including domains, seed environment specifications, evolution trajectories, and task generation parameters. Appendix~\ref{app:agent-results} details the evaluation setup including specific model versions, temperature settings, number of runs per task, and the user simulator configuration. Full prompt specifications are provided in Appendix~\ref{app:prompts-implementation}.
    \item[] Guidelines:
    \begin{itemize}
        \item The answer \answerNA{} means that the paper does not include experiments.
        \item The experimental setting should be presented in the core of the paper to a level of detail that is necessary to appreciate the results and make sense of them.
        \item The full details can be provided either with the code, in appendix, or as supplemental material.
    \end{itemize}

\item {\bf Experiment statistical significance}
    \item[] Question: Does the paper report error bars suitably and correctly defined or other appropriate information about the statistical significance of the experiments?
    \item[] Answer: \answerNo{} 
    \item[] Justification: For all LLM inference calls (both environment generation and downstream agent evaluation), we set temperature=0 to ensure deterministic outputs. The downstream agent results serve as diagnostic evidence rather than leaderboard rankings, and are aggregated over large-scale evaluation across environments with two domains, where the systematic trends across evolution primitives reflect structured environment differences rather than sampling noise.
    
    \item[] Guidelines:
    \begin{itemize}
        \item The answer \answerNA{} means that the paper does not include experiments.
        \item The authors should answer \answerYes{} if the results are accompanied by error bars, confidence intervals, or statistical significance tests, at least for the experiments that support the main claims of the paper.
        \item The factors of variability that the error bars are capturing should be clearly stated (for example, train/test split, initialization, random drawing of some parameter, or overall run with given experimental conditions).
        \item The method for calculating the error bars should be explained (closed form formula, call to a library function, bootstrap, etc.)
        \item The assumptions made should be given (e.g., Normally distributed errors).
        \item It should be clear whether the error bar is the standard deviation or the standard error of the mean.
        \item It is OK to report 1-sigma error bars, but one should state it. The authors should preferably report a 2-sigma error bar than state that they have a 96\% CI, if the hypothesis of Normality of errors is not verified.
        \item For asymmetric distributions, the authors should be careful not to show in tables or figures symmetric error bars that would yield results that are out of range (e.g., negative error rates).
        \item If error bars are reported in tables or plots, the authors should explain in the text how they were calculated and reference the corresponding figures or tables in the text.
    \end{itemize}

\item {\bf Experiments compute resources}
    \item[] Question: For each experiment, does the paper provide sufficient information on the computer resources (type of compute workers, memory, time of execution) needed to reproduce the experiments?
    \item[] Answer: \answerYes{} 
    \item[] Justification: All experiments use commercial LLM API calls (GPT-5, Claude-Opus-4.5, Gemini-2.5-Pro, DeepSeek-V3.2, Qwen3-235B) via LiteLLM. We report estimated API costs per task in Table~\ref{table:episode_results} and Table~\ref{tab:airline_trajectory}. No GPU training is involved; all computation is inference-based through cloud API providers.

    \item[] Guidelines:
    \begin{itemize}
        \item The answer \answerNA{} means that the paper does not include experiments.
        \item The paper should indicate the type of compute workers CPU or GPU, internal cluster, or cloud provider, including relevant memory and storage.
        \item The paper should provide the amount of compute required for each of the individual experimental runs as well as estimate the total compute. 
        \item The paper should disclose whether the full research project required more compute than the experiments reported in the paper (e.g., preliminary or failed experiments that didn't make it into the paper). 
    \end{itemize}
    
\item {\bf Code of ethics}
    \item[] Question: Does the research conducted in the paper conform, in every respect, with the NeurIPS Code of Ethics \url{https://neurips.cc/public/EthicsGuidelines}?
    \item[] Answer: \answerYes{} 
    \item[] Justification: We fully adhere to the NeurIPS Code of Ethics.
    \item[] Guidelines: 
    \begin{itemize}
        \item The answer \answerNA{} means that the authors have not reviewed the NeurIPS Code of Ethics.
        \item If the authors answer \answerNo, they should explain the special circumstances that require a deviation from the Code of Ethics.
        \item The authors should make sure to preserve anonymity (e.g., if there is a special consideration due to laws or regulations in their jurisdiction).
    \end{itemize}

\item {\bf Broader impacts}
    \item[] Question: Does the paper discuss both potential positive societal impacts and negative societal impacts of the work performed?
    \item[] Answer: \answerYes{}{} 
    \item[] Justification: See Appendix \ref{app:impact}.
    \item[] Guidelines:
    \begin{itemize}
        \item The answer \answerNA{} means that there is no societal impact of the work performed.
        \item If the authors answer \answerNA{} or \answerNo, they should explain why their work has no societal impact or why the paper does not address societal impact.
        \item Examples of negative societal impacts include potential malicious or unintended uses (e.g., disinformation, generating fake profiles, surveillance), fairness considerations (e.g., deployment of technologies that could make decisions that unfairly impact specific groups), privacy considerations, and security considerations.
        \item The conference expects that many papers will be foundational research and not tied to particular applications, let alone deployments. However, if there is a direct path to any negative applications, the authors should point it out. For example, it is legitimate to point out that an improvement in the quality of generative models could be used to generate Deepfakes for disinformation. On the other hand, it is not needed to point out that a generic algorithm for optimizing neural networks could enable people to train models that generate Deepfakes faster.
        \item The authors should consider possible harms that could arise when the technology is being used as intended and functioning correctly, harms that could arise when the technology is being used as intended but gives incorrect results, and harms following from (intentional or unintentional) misuse of the technology.
        \item If there are negative societal impacts, the authors could also discuss possible mitigation strategies (e.g., gated release of models, providing defenses in addition to attacks, mechanisms for monitoring misuse, mechanisms to monitor how a system learns from feedback over time, improving the efficiency and accessibility of ML).
    \end{itemize}
    
\item {\bf Safeguards}
    \item[] Question: Does the paper describe safeguards that have been put in place for responsible release of data or models that have a high risk for misuse (e.g., pre-trained language models, image generators, or scraped datasets)?
    \item[] Answer: \answerNA{} 
    \item[] Justification: The paper poses no such risks.
    \item[] Guidelines:
    \begin{itemize}
        \item The answer \answerNA{} means that the paper poses no such risks.
        \item Released models that have a high risk for misuse or dual-use should be released with necessary safeguards to allow for controlled use of the model, for example by requiring that users adhere to usage guidelines or restrictions to access the model or implementing safety filters. 
        \item Datasets that have been scraped from the Internet could pose safety risks. The authors should describe how they avoided releasing unsafe images.
        \item We recognize that providing effective safeguards is challenging, and many papers do not require this, but we encourage authors to take this into account and make a best faith effort.
    \end{itemize}

\item {\bf Licenses for existing assets}
    \item[] Question: Are the creators or original owners of assets (e.g., code, data, models), used in the paper, properly credited and are the license and terms of use explicitly mentioned and properly respected?
    \item[] Answer: \answerYes{} 
    \item[] Justification: The paper properly credits the creators or original owners of the assets used, such as code, data, and models. We also have correctly cited the relevant literature.
    \item[] Guidelines:
    \begin{itemize}
        \item The answer \answerNA{} means that the paper does not use existing assets.
        \item The authors should cite the original paper that produced the code package or dataset.
        \item The authors should state which version of the asset is used and, if possible, include a URL.
        \item The name of the license (e.g., CC-BY 4.0) should be included for each asset.
        \item For scraped data from a particular source (e.g., website), the copyright and terms of service of that source should be provided.
        \item If assets are released, the license, copyright information, and terms of use in the package should be provided. For popular datasets, \url{paperswithcode.com/datasets} has curated licenses for some datasets. Their licensing guide can help determine the license of a dataset.
        \item For existing datasets that are re-packaged, both the original license and the license of the derived asset (if it has changed) should be provided.
        \item If this information is not available online, the authors are encouraged to reach out to the asset's creators.
    \end{itemize}

\item {\bf New assets}
    \item[] Question: Are new assets introduced in the paper well documented and is the documentation provided alongside the assets?
    \item[] Answer: \answerYes{} 
    \item[] Justification: The paper introduces a new benchmark framework (ProEvolve) comprising 200 e-commerce environment versions, 20 airline environment versions, and 3,300 task sandboxes. The benchmark construction pipeline, generated environments, and evaluation code will be publicly released upon acceptance under an open-source license. The paper documents the generation methodology (Sections~4.1--4.3), validation criteria (Section~5.2), and full prompt specifications (Appendix~\ref{app:prompts-implementation}).
    \item[] Guidelines:
    \begin{itemize}
        \item The answer \answerNA{} means that the paper does not release new assets.
        \item Researchers should communicate the details of the dataset\slash code\slash model as part of their submissions via structured templates. This includes details about training, license, limitations, etc. 
        \item The paper should discuss whether and how consent was obtained from people whose asset is used.
        \item At submission time, remember to anonymize your assets (if applicable). You can either create an anonymized URL or include an anonymized zip file.
    \end{itemize}

\item {\bf Crowdsourcing and research with human subjects}
    \item[] Question: For crowdsourcing experiments and research with human subjects, does the paper include the full text of instructions given to participants and screenshots, if applicable, as well as details about compensation (if any)? 
    \item[] Answer: \answerNA{} 
    \item[] Justification: The paper does not involve crowdsourcing nor research with human subject. 
    \item[] Guidelines:
    \begin{itemize}
        \item The answer \answerNA{} means that the paper does not involve crowdsourcing nor research with human subjects.
        \item Including this information in the supplemental material is fine, but if the main contribution of the paper involves human subjects, then as much detail as possible should be included in the main paper. 
        \item According to the NeurIPS Code of Ethics, workers involved in data collection, curation, or other labor should be paid at least the minimum wage in the country of the data collector. 
    \end{itemize}

\item {\bf Institutional review board (IRB) approvals or equivalent for research with human subjects}
    \item[] Question: Does the paper describe potential risks incurred by study participants, whether such risks were disclosed to the subjects, and whether Institutional Review Board (IRB) approvals (or an equivalent approval/review based on the requirements of your country or institution) were obtained?
    \item[] Answer: \answerNA{} 
    \item[] Justification:  The paper does not involve crowdsourcing nor research with human subject
    \item[] Guidelines:
    \begin{itemize}
        \item The answer \answerNA{} means that the paper does not involve crowdsourcing nor research with human subjects.
        \item Depending on the country in which research is conducted, IRB approval (or equivalent) may be required for any human subjects research. If you obtained IRB approval, you should clearly state this in the paper. 
        \item We recognize that the procedures for this may vary significantly between institutions and locations, and we expect authors to adhere to the NeurIPS Code of Ethics and the guidelines for their institution. 
        \item For initial submissions, do not include any information that would break anonymity (if applicable), such as the institution conducting the review.
    \end{itemize}

\item {\bf Declaration of LLM usage}
    \item[] Question: Does the paper describe the usage of LLMs if it is an important, original, or non-standard component of the core methods in this research? Note that if the LLM is used only for writing, editing, or formatting purposes and does \emph{not} impact the core methodology, scientific rigor, or originality of the research, declaration is not required.
    \item[] Answer: \answerYes{} 
    \item[] Justification: LLMs are a core component of the proposed framework. Specifically: (1) an LLM agent proposes evolution intents and generates graph transformations (Section~4.2, Phase~I); (2) an LLM coding agent instantiates transformed graphs into executable code (Section~4.2, Phase~II); (3) an LLM generates unit tests for validation (Appendix~\ref{app:prompts-implementation}); (4) an LLM-as-a-Judge evaluates semantic fidelity of each modification (Section~5.2); and (5) an LLM-powered user simulator conducts agent evaluation (Section~4.4). All prompt templates are provided in Appendix~\ref{app:prompts-implementation}.
    
    \item[] Guidelines:
    \begin{itemize}
        \item The answer \answerNA{} means that the core method development in this research does not involve LLMs as any important, original, or non-standard components.
        \item Please refer to our LLM policy in the NeurIPS handbook for what should or should not be described.
    \end{itemize}

\end{enumerate}

\end{document}